\documentclass[journal]{IEEEtran}
\usepackage{graphicx}
\usepackage{amsmath,amssymb} % define this before the line numbering.
\usepackage{color}
\usepackage{epsfig}
\usepackage{soul}
\usepackage{mathtools}
\usepackage[font=small,skip=-1pt]{caption}
\usepackage[font=small]{subfig}
\usepackage{xcolor, colortbl}
\usepackage{multirow}
\usepackage{tabularx}
\usepackage{algorithm}
\usepackage{algpseudocode}
\usepackage{algorithmicx}
\usepackage{microtype}
\usepackage{booktabs}
\usepackage{enumitem}
\usepackage{bbm}
\usepackage[super]{nth}
\newcolumntype{Y}{>{\centering\arraybackslash}X}
\ifCLASSINFOpdf
\else
\fi
\begin{document}
	
\setlength{\belowcaptionskip}{-5pt}
\setlength{\abovedisplayskip}{5pt}
\setlength{\belowdisplayskip}{5pt}
\title{Estimation of Camera Response Function using Prediction Consistency and Gradual Refinement with an Extension to Deep Learning}

\author{Aashish Sharma,
        Robby T. Tan,~\IEEEmembership{Member,~IEEE,}
        and~Loong-Fah Cheong,~\IEEEmembership{Member,~IEEE}% <-this % stops a space
\thanks{A. Sharma, R. T. Tan and L.F. Cheong are with the Department of Electrical and Computer Engineering, National University of Singapore, Singapore. e-mail: (aashish.sharma@u.nus.edu, eleclf@nus.edu.sg, robby.tan@nus.edu.sg).}% <-this % stops a space
%\thanks{R. T. Tan is also with Yale-NUS College, Singapore.}% <-this % stops a space
}


\markboth{Journal of \LaTeX\ Class Files,~Vol.~14, No.~8, August~2015}%
{Shell \MakeLowercase{\textit{et al.}}: Bare Demo of IEEEtran.cls for IEEE Journals}



\maketitle

\begin{abstract}
Most existing methods for CRF estimation from a single image fail to handle general real images. 
For instance, EdgeCRF~\cite{lin2004radiometric} based on colour patches extracted from edges works effectively only when the presence of noise is insignificant, which is not the case for many real images; and, CRFNet~\cite{li2017crf}, a recent method based on fully supervised deep learning works only for the CRFs that are in the training data, and hence fail to deal with other possible CRFs beyond the training data. 
To address these problems, we introduce a non-deep-learning method using prediction consistency and gradual refinement. 
First, we rely more on the patches of the input image that provide more consistent predictions. If the predictions from a patch are more consistent, it means that the patch is likely to be less affected by noise or any inferior colour combinations, and hence, it can be more reliable for CRF estimation.
Second, we employ a gradual refinement scheme in which we start from a simple CRF model to generate a result which is more robust to noise but less accurate, and then we gradually increase the model's complexity to improve the result. This is because a simple model, while being less accurate, overfits less to noise than a complex model does. Our experiments show that our method outperforms the existing single-image methods for daytime and nighttime real images.
We further propose a more efficient deep learning extension that performs test-time training (based on unsupervised losses) on the test input image. This provides our method better generalization performance than CRFNet~\cite{li2017crf} making it more practically applicable for CRF estimation for general real images. 
\end{abstract}


% \fi
\IEEEpeerreviewmaketitle

\section{Introduction}
Due to the non-linearity of the Camera Response Function (CRF), in most cameras, the camera irradiance has a non-linear correlation to the image intensities. 
In the camera imaging pipeline, CRF is one of the main components and is intentionally designed to be non-linear to create more aesthetic effects and perform dynamic range compression~\cite{kim2008radiometric}.
Estimating the CRF enables us to linearise the image intensities which is critical for many computer vision algorithms such as shape from shading~\cite{zhang1999shape,nayar1991shape}, colour constancy~\cite{finlayson2001color,tan2004color}, photometric stereo~\cite{shi2010self,shi2014photometric}, specular removal \cite{tan2005specular}, 
%intrinsic image decomposition~\cite{weiss2001deriving}, 
shadow removal~\cite{finlayson2009entropy}, low-light image enhancement~\cite{ying2017new}, etc.

In the early approaches of CRF estimation, some methods use multiple images of a static scene taken with different exposures~\cite{debevec2008recovering,mitsunaga1999radiometric,mann2000comparametric,shi2014photometric}. 
To improve the practicality, some recent methods utilise a single image. This, however, is considerably more challenging than using multiple images. 
To accomplish this, the methods require additional constraints, such as: 
the symmetric nature of the irradiance noise profile~\cite{matsushita2007radiometric},
or  
the non-linear distributions of the pixel intensities in the RGB space for pixels around edges~\cite{lin2004radiometric}.
The former constraint used by~\cite{matsushita2007radiometric} is applicable only to images that have a considerable large range of intensity values, which is not the case for many real images. 
The latter constraint used by~\cite{lin2004radiometric} is more applicable to real images, since it only relies on non-uniform coloured patches in RGB images. 
Unfortunately, the existing methods that use this constraint are erroneous when noise is present in the input image, which is the case for many real daytime and nighttime images since noise is usually unavoidable~\cite{wei2020physics}. 
Recently, a method \cite{li2017crf} based on fully supervised deep learning is proposed. It also requires only a single image to estimate the CRF and can work robustly on real images. However, it suffers from the generalization problem which means that it can only work effectively for CRFs that are represented in the training data.

\begin{figure}[t!]
	\captionsetup[subfloat]{labelformat=empty}
	\captionsetup[subfloat]{farskip=4pt}
	\begin{center}
		\subfloat{\includegraphics[width=0.245\textwidth]{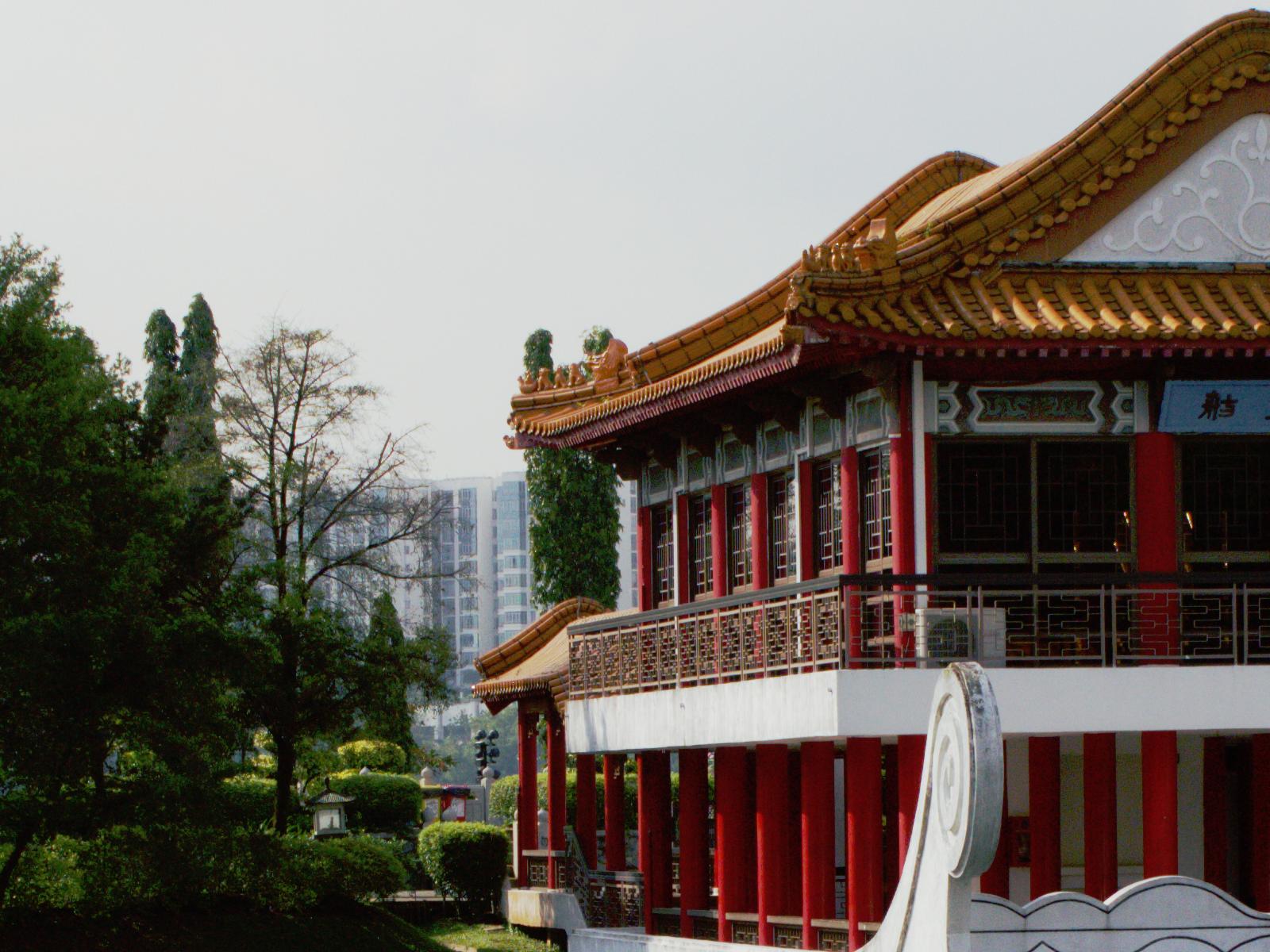}}\hfill
		\subfloat{\includegraphics[width=0.24\textwidth]{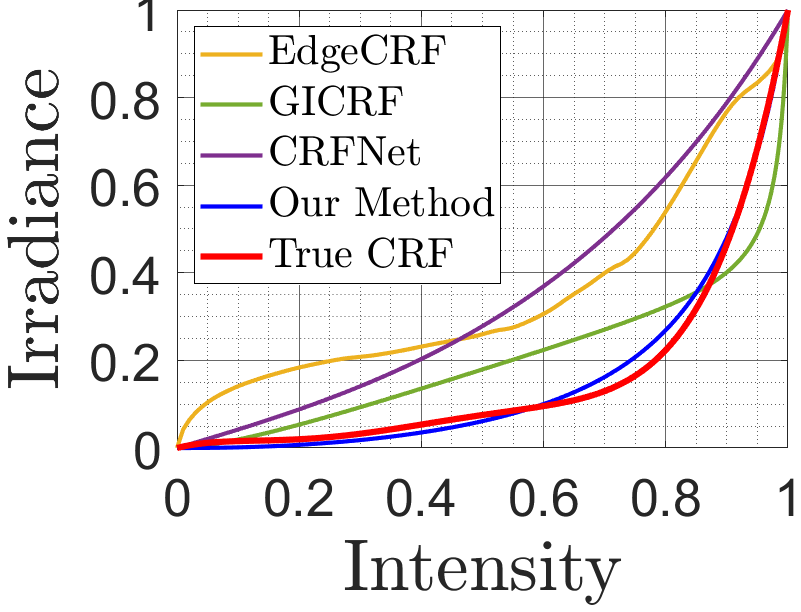}}\hfill\\
				
		\subfloat[Input Image]{\includegraphics[width=0.245\textwidth]{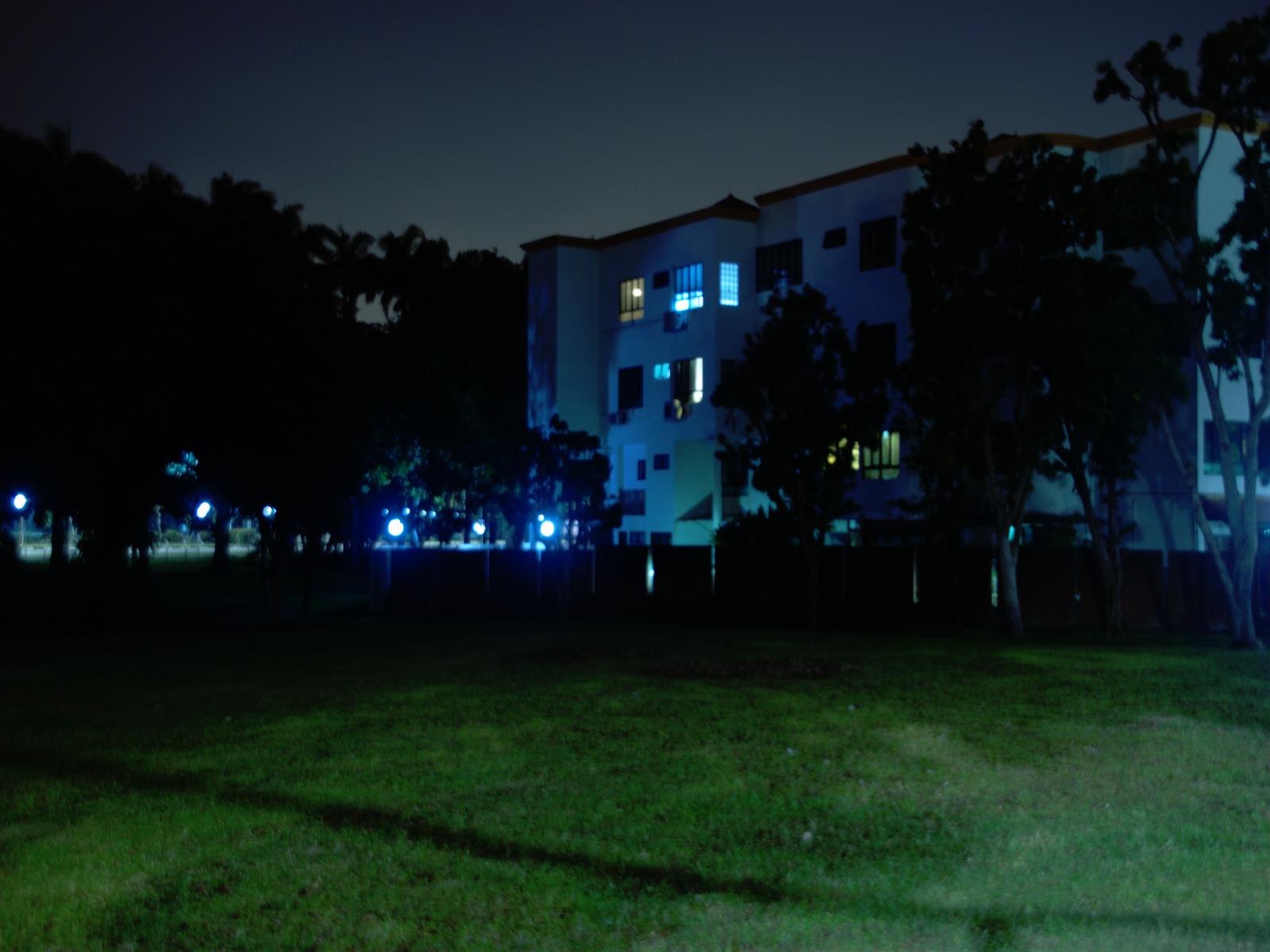}}\hfill
		\subfloat[Predicted CRF]{\includegraphics[width=0.24\textwidth]{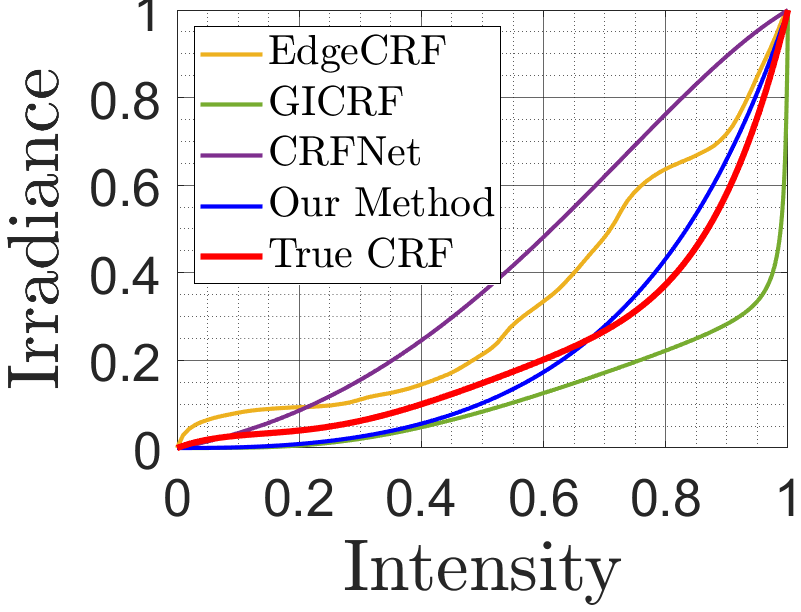}}\hfill
	\end{center}
	\caption{Results of our CRF estimation method in comparison with existing methods EdgeCRF~\cite{lin2004radiometric}, GICRF~\cite{ng2007using} and CRFNet~\cite{li2017crf} on daytime and nighttime images. We can observe that our predicted CRF is closest to the true CRF. See Fig.~\ref{figure_trailer2} for additional examples.}\label{figure_trailer}
\end{figure}

\begin{figure*}[t!]
	\captionsetup[subfloat]{labelformat=empty}
	\captionsetup[subfloat]{farskip=4pt}
	\begin{center}
		\subfloat{\includegraphics[width=0.245\textwidth]{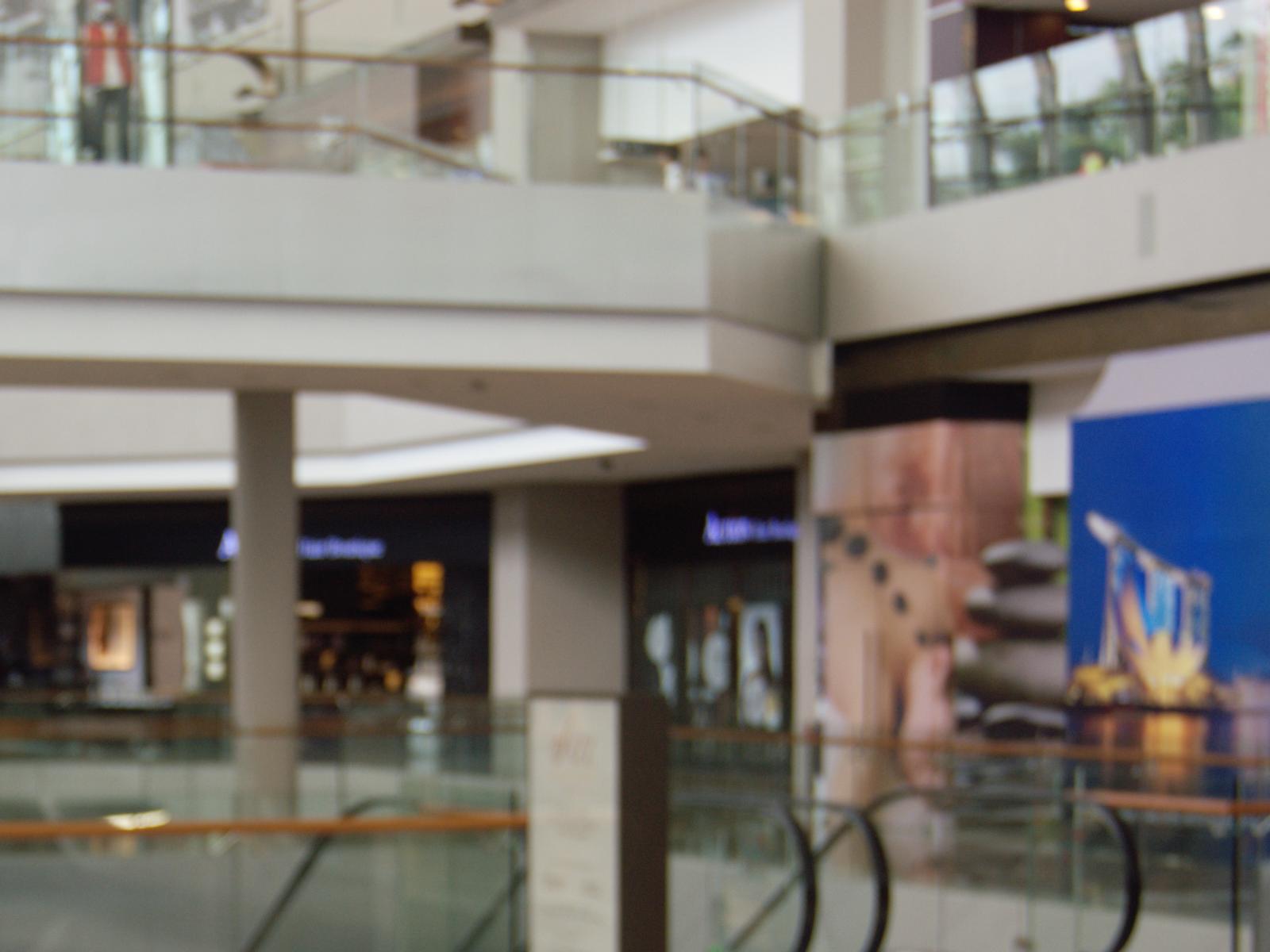}}\hfill
		\subfloat{\includegraphics[width=0.24\textwidth]{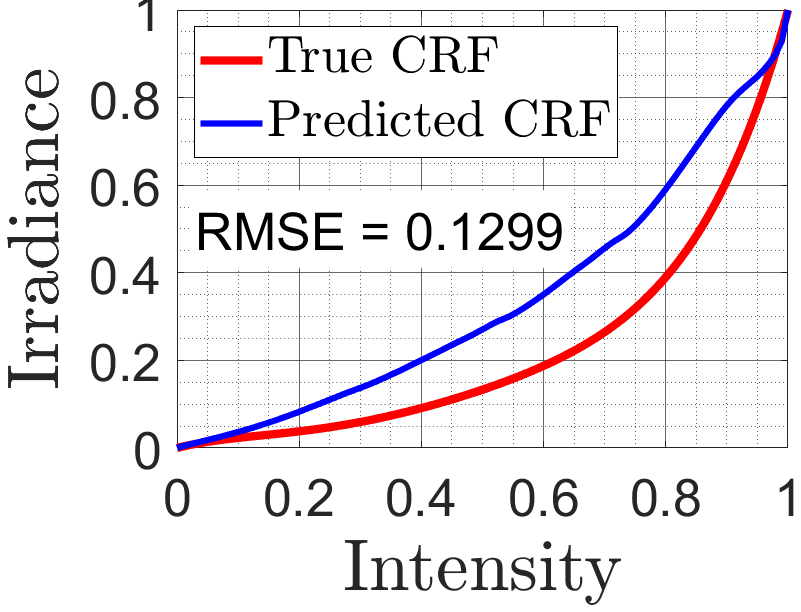}}\hfill
		\subfloat{\includegraphics[width=0.24\textwidth]{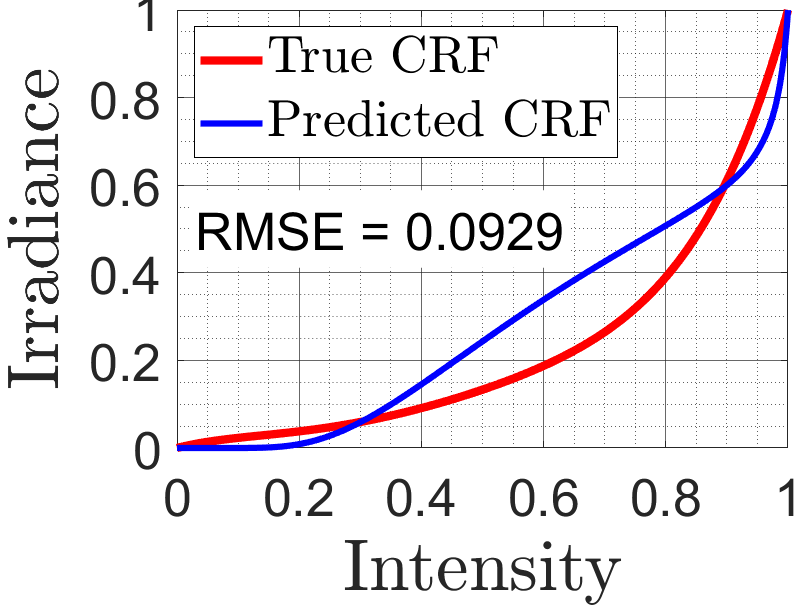}}\hfill
		\subfloat{\includegraphics[width=0.24\textwidth]{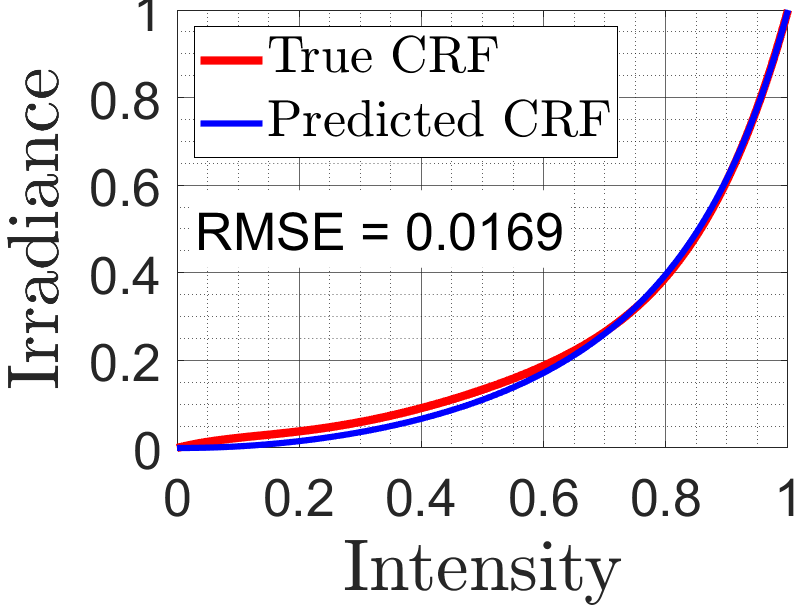}}\hfill\\
		
		\subfloat[Input Image]{\includegraphics[width=0.245\textwidth]{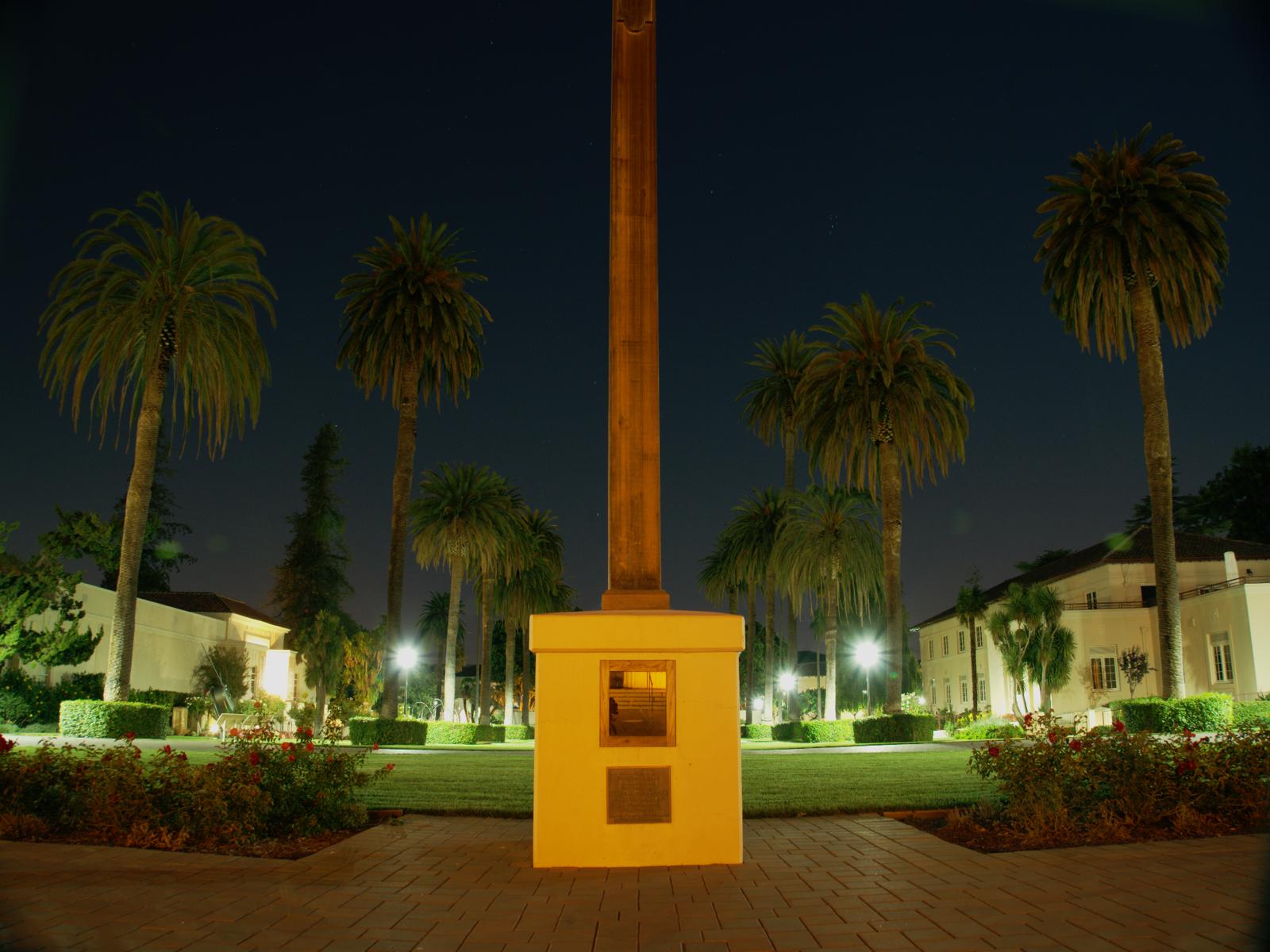}}\hfill
		\subfloat[EdgeCRF~\cite{lin2004radiometric}]{\includegraphics[width=0.24\textwidth]{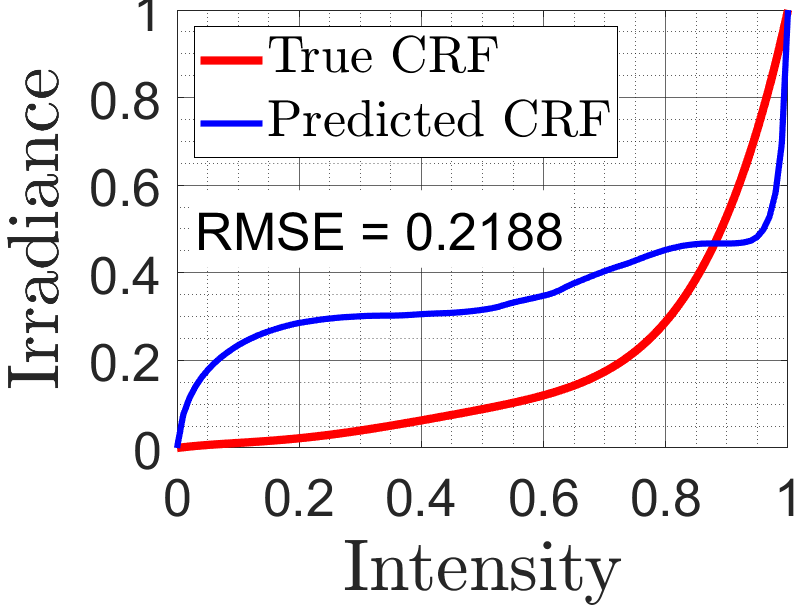}}\hfill
		\subfloat[GICRF~\cite{ng2007using}]{\includegraphics[width=0.24\textwidth]{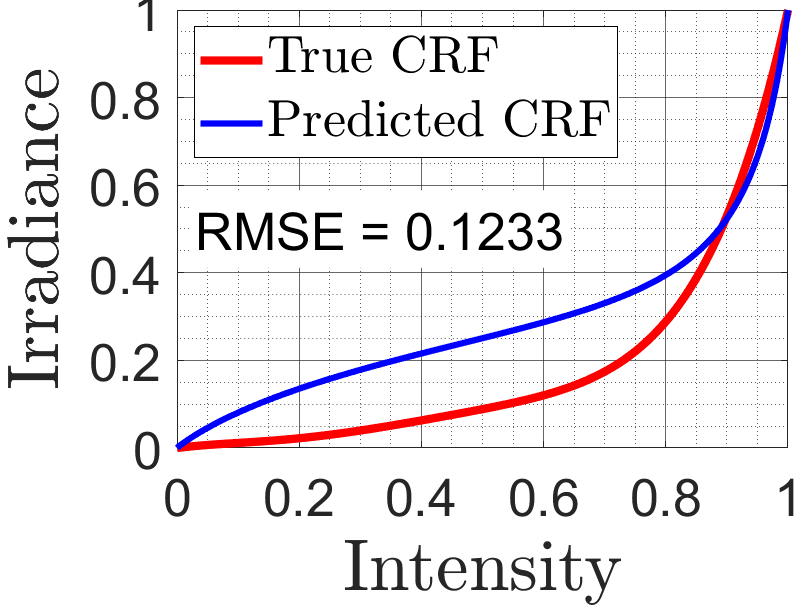}}\hfill
		\subfloat[\textbf{Our Method}]{\includegraphics[width=0.24\textwidth]{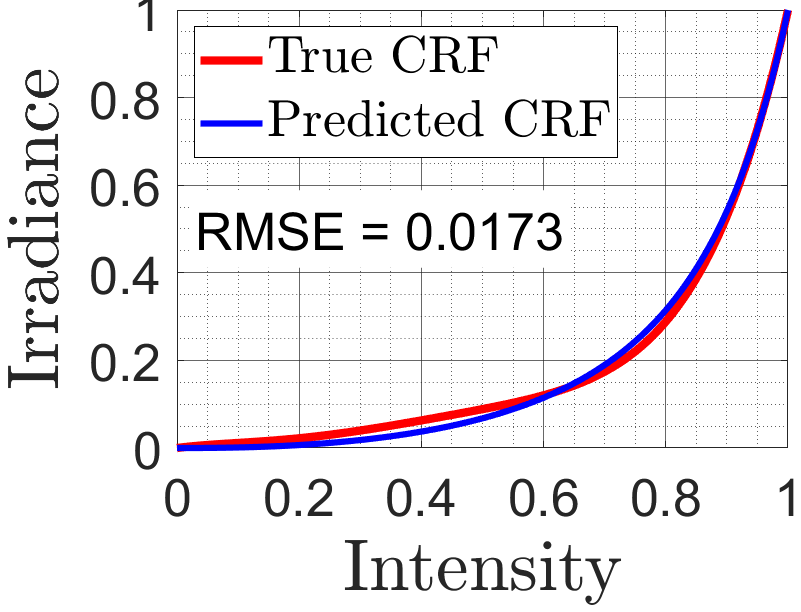}}\hfill\\
	\end{center}
	\caption{Results of our single-image CRF estimation method in comparison with those of two existing methods, EdgeCRF~\cite{lin2004radiometric} and GICRF~\cite{ng2007using}, for both daytime and nighttime image. RMSE is the Root Mean Squares Error between the predicted and true CRF.}\label{figure_trailer2}
\end{figure*}

To address the problem of single-image CRF estimation for general real images, we introduce a non-deep-learning method (published as~\cite{sharma2020single}) based on the ideas of prediction consistency and gradual refinement. 
First, unlike the existing methods, our method relies more on patches that provide more consistent CRF predictions. If a patch provides more consistent CRF predictions, it is likely to be less affected by noise or any inferior colour combinations, and hence, can be more reliable to estimate the CRF. We put more weight on more reliable patches, which renders more accurate CRF estimation.
Second, unlike the existing methods, our method employs a gradual refinement scheme.  
%
%CRF estimation is generally carried out by optimising a CRF model that can linearise the non-linear pixel distributions with minimum error~\cite{lin2004radiometric}. 
%%
%There are various models that can be employed to represent CRFs  . 
%%
%Each model is defined by a set of coefficients, hence the problem of finding the CRF is equivalent to optimizing the coefficients of the CRF model.
%
The problem of finding the CRF is equivalent to optimizing the coefficients of the CRF model (e.g.,~\cite{grossberg2004modeling,shi2010self,lee2012radiometric,ng2007using}). 
A model with more coefficients is more accurate, but is also more prone to overfit to noise. 
Therefore, to gain robustness to noise, our method optimises a simple model first (i.e., one coefficient), and then optimises a more complex model that improves our CRF result.
During the refinement, the CRF result in the current stage is constrained to remain near the CRF result obtained in the previous stage. 
Figs.~\ref{figure_trailer} and \ref{figure_trailer2} show our method's better performance for both daytime and nighttime real images. 

This paper is based on our previous paper~\cite{sharma2020single}, where we first introduce the ideas of prediction consistency and gradual refinement to estimate the CRF from a single image. 
In this paper, we provide more details, analysis and applications of our method~\cite{sharma2020single} which will help our readers to better understand our method.
In addition,  we propose a more efficient deep learning extension where different from our original method~\cite{sharma2020single}, we use a network to estimate the CRF from a single image. In contrast to the existing approaches~\cite{li2017crf} that use fully supervised learning and are more prone to the generalization problem, we use test-time training based on unsupervised losses. This improves our generalisation performance and makes our method more practically applicable to real images compared to the existing methods. 

To summarise, our main contributions are as follows:
\begin{itemize}[noitemsep,topsep=1pt]
	\item Unlike existing methods, our method is designed to deal with noise, making it more applicable to general real images. For achieving this, we propose the ideas of prediction consistency and gradual refinement.
	%first, we found that different patches are unequal in terms of noise and how much they carry the CRF information (or degree of non-linearity). For instance, in one input image, patches taken from darker regions are more affected by noise compared to patches taken from well-lit regions; on top of this, in a patch, certain colour combinations carry more CRF information compared to other colour combinations. Second, unlike  \cite{lin2004radiometric}, we found non-edge patches can be used to estimate a CRF. Third, we found noise can significantly alter the degree of non-linearity in a patch, which is critical for a robust method to consider. All these findings are new and important for CRF estimation.
	%
	%\item 
	Prediction consistency is used to compute the reliability of a patch in estimating CRFs. It is based on the consistency among a patch's CRF predictions. More consistency implies more reliability. 
	%Our method puts more emphasis on the more reliable patches that generates more accurate results. 
	%
	%\item 
	Under our gradual refinement scheme, we start from a simple model to obtain results that are robust to noise but less accurate, and then we gradually increase the model's complexity to improve the results. 
	%We show that our refinement scheme generates better results than those of  one-attempt estimation methods. 
	%The idea of gradual refinement is, in various forms, already used in computer vision algorithms, however to our knowledge, it is the first time it is designed specifically for single-image CRF estimation.
	To our knowledge, both prediction consistency and gradual refinement are new in single-image CRF estimation. 
	\item We introduce a more efficient deep learning extension by proposing  a network to estimate the CRF. Unlike the existing baseline methods (e.g.~\cite{li2017crf}), we perform test-time training on the test input image using unsupervised losses that provides us better generalization performance.
\end{itemize} 
Our method has practical applications such as nighttime visibility enhancement that improves the perceptual quality of the input nighttime image and also improves the performance of vision tasks such as object detection. 

\begin{figure*}[t!]
	%\captionsetup[subfloat]{labelformat=empty}
	\captionsetup[subfloat]{farskip=1pt}
	\begin{center}
		\subfloat[Input Image $\mathbf{I}$]{\label{figure_edge_dists_img}\includegraphics[width=0.25\textwidth]{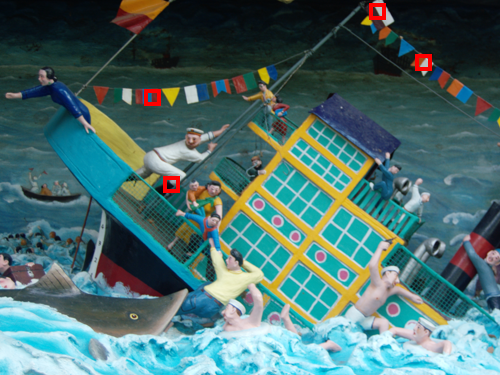}}\hfill
		\subfloat[Patches]{\label{figure_edge_dists_patches}\includegraphics[width=0.20\textwidth]{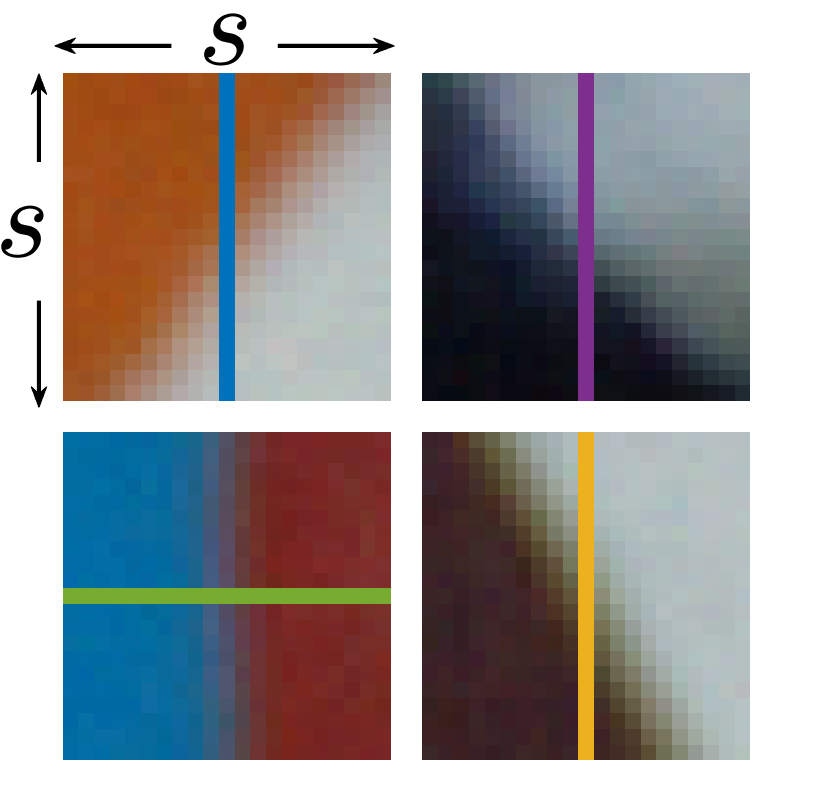}}\hfill
		\subfloat[Pixel Distributions]{\label{figure_edge_dists_dists}\includegraphics[width=0.26\textwidth]{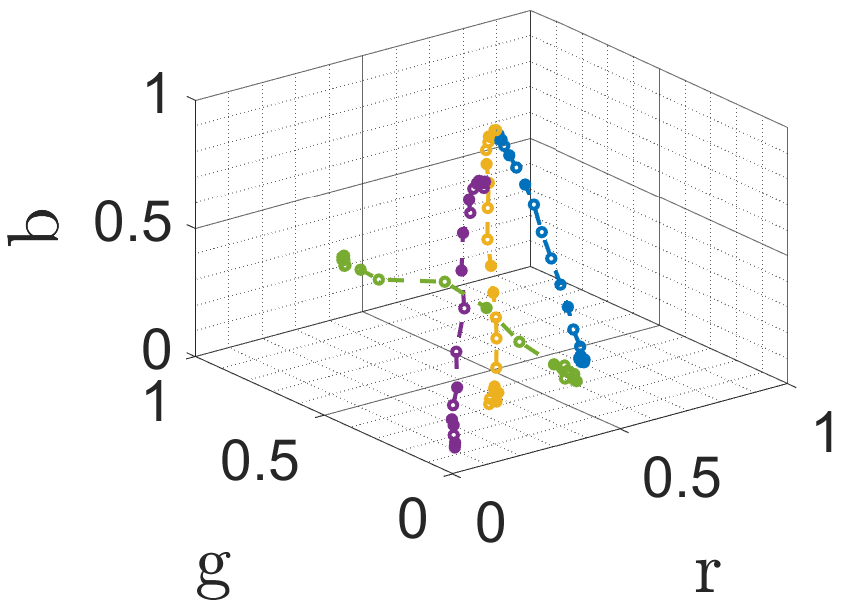}}\hfill
		\subfloat[True CRF $g$]{\label{figure_edge_dists_crf}\includegraphics[width=0.24\textwidth]{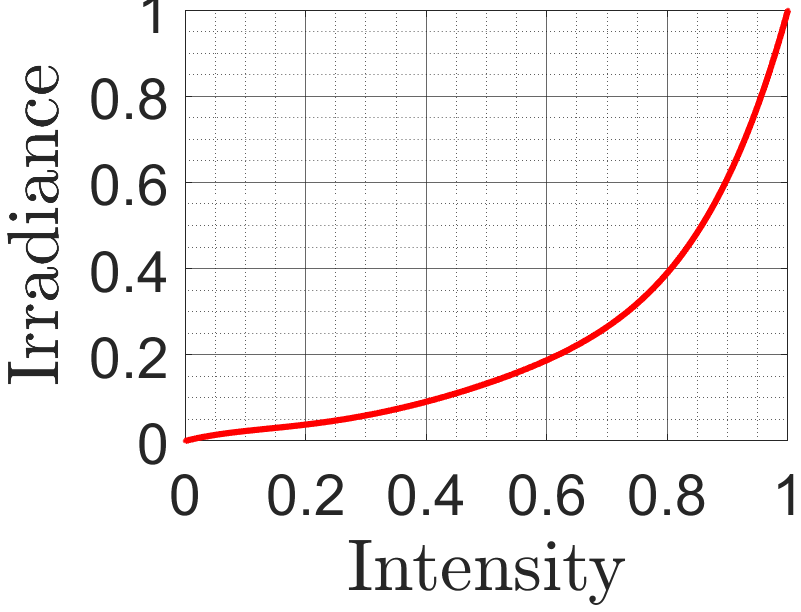}}\hfill
	\end{center}
	\caption{(a) Input image $\mathbf{I}$. (b) Patches around edges in the image $\mathbf{I}$. Each patch is of size $s\times s$. The coloured line in a patch shows a pixel distribution scanned horizontally or vertically from that patch. (c) Pixel distributions of the patches plotted in the RGB space. (d) True CRF $g$.}
	\label{figure_edge_dists}
\end{figure*}

\section{Related Work}
Several methods have been proposed for CRF estimation. Many previous works rely on a sequence of images of the same scene taken with different exposure times~\cite{debevec2008recovering,mitsunaga1999radiometric,mann2000comparametric} or varying illumination~\cite{shi2010self,kim2008radiometric}. 
In addition to the requirement of multiple images, these methods also assume that the camera position is fixed. 
Some other methods~\cite{kim2008robust,litvinov2005addressing,mann2001quantigraphic,park2016efficient,diaz2011radiometric} have relaxed the fixed camera condition by performing image alignment. 
In contrast to these multi-image methods, our method requires a single image, and hence, is more applicable. 

Lin et al.~\cite{lin2004radiometric} are the first to demonstrate that the CRF can be recovered from a single colour image. 
Their method is based on the property that pixels around color boundaries in the camera irradiance form linear distributions in the RGB space. However, due to the non-linear CRF, the pixel distributions become non-linear for the intensity image in the RGB space. 
Thus the CRF is recovered by finding a CRF model that can linearise the distributions with minimum error. 
Later, the method is extended to work on grayscale images~\cite{lin2005determining}.
The ideas of these two methods are elegant, yet unfortunately they are prone to suffer from noise that is commonly present in real images. 
Matsushita and Lin~\cite{matsushita2007radiometric} use noise profile for CRF estimation, based on the observation that the profile is symmetric in the irradiance domain but due to the non-linear CRF, it becomes asymmetric in the intensity domain. 
However, this method does not work well if there is insufficient data available for generating the noise profile. 
Li et al.~\cite{li2017radiometric} propose to use the low-rank structure of skin pixels to recover the CRF from an image that contains a human face. This method, therefore, is limited to work on images that consist of human faces.
Recently, Li and Peers~\cite{li2017crf} propose a network to estimate the CRF from a single image by predicting 11 coefficients of the basis functions model~\cite{grossberg2004modeling}, for which they use the 201 CRFs from~\cite{grossberg2004modeling} during training. Kao et al. further propose~\cite{kao2018progressive}, a stacked version of~\cite{li2017crf}, which shares a similar idea of iterative refinement with our method. Both \cite{li2017crf} and \cite{kao2018progressive} are supervised learning-based methods and suffer from the generalization problem. This means they cannot work properly for estimating the CRFs that are not represented in the training set. In contrast, our method is not a learning-based method, and hence, does not suffer from the generalization problem.

Similar to Lin et al.~\cite{lin2004radiometric}, our method uses the non-linear pixel distributions to estimate the CRF from a single image. However, in contrast to the method, our method uses a consistency based metric to take into account the reliability of the patches (containing the distributions) for estimating the CRF. Additionally, unlike the one-attempt estimation approach of their method, our method uses a gradual refinement scheme that provides more robust and accurate CRF results.

\section{Proposed Method}
The camera imaging pipeline can be represented by~\cite{kim2012new}:
\begin{equation}\label{eq_cam_model}
\mathbf{I} = f(T(\mathbf{E})), 
\end{equation}
where $\mathbf{E}$ is the  camera irradiance in the RGB colour channels, $T$ is a linear operator for colour transformations (for e.g., white balance), $f$ is a non-linear function representing the CRF, and $\mathbf{I}$ is the RGB  image outputted by the camera.

Given an image $\mathbf{I}$ (normalised in $[0,1]$) as input, our goal is to estimate the inverse CRF $g=f^{-1}$, such that the image $g(\mathbf{I})$ becomes linearly related to the camera irradiance $\mathbf{E}$. Figs.~\ref{figure_edge_dists_img} and \ref{figure_edge_dists_crf} shows an example of $\mathbf{I}$ and $g=f^{-1}$ respectively.

\begin{figure*}[t!]
	\captionsetup[subfloat]{labelformat=empty}
	\captionsetup[subfloat]{farskip=1pt}
	\begin{center}
		\begin{center}
			\subfloat[Patches]{\label{figure_synth_ill_1_cp}\includegraphics[width=0.21\textwidth]{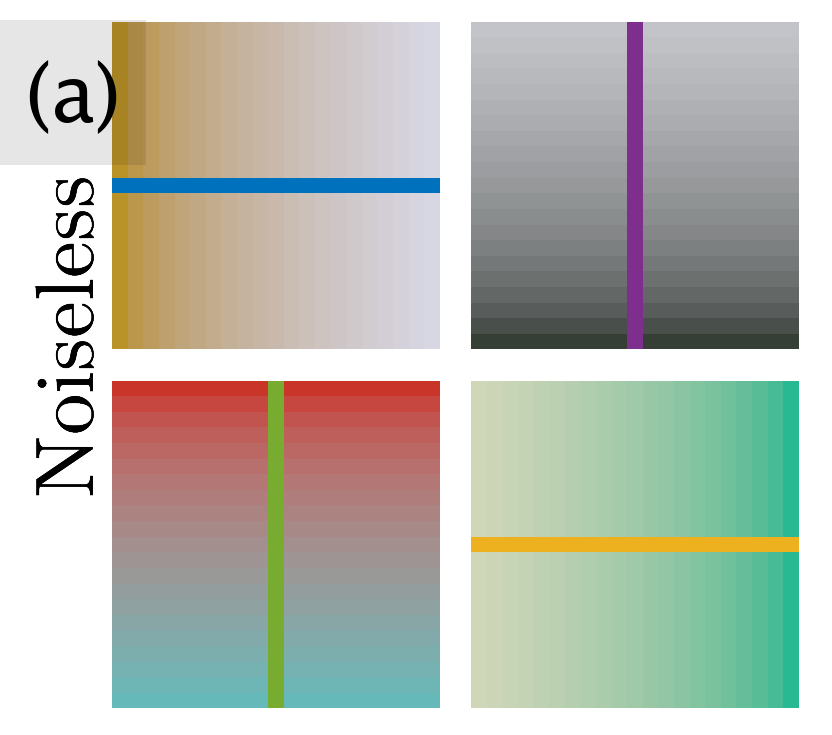}}\hfill
			\subfloat[Pixel Distributions]{\label{figure_synth_ill_1_cpdists}\includegraphics[width=0.24\textwidth]{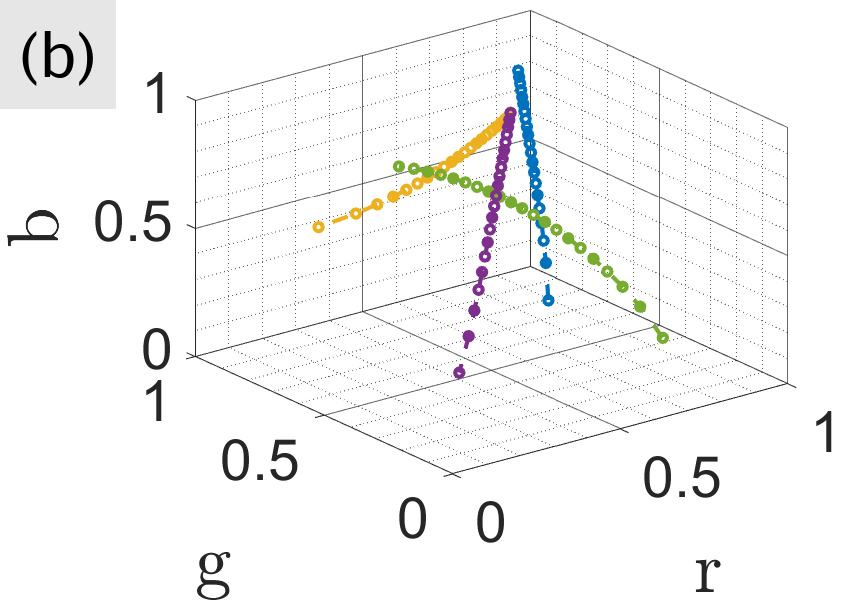}}\hfill
			\subfloat[Profiles (without $N$)]{\label{figure_synth_ill_1_cpprofs}\includegraphics[width=0.24\textwidth]{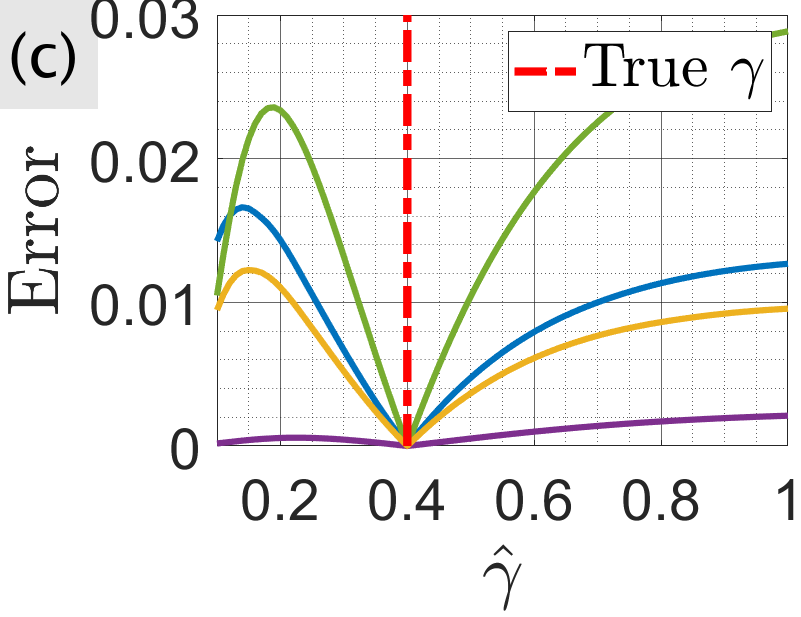}}\hfill
			\subfloat[Profiles (with $N$)]{\label{figure_synth_ill_1_cpprofsN}\includegraphics[width=0.24\textwidth]{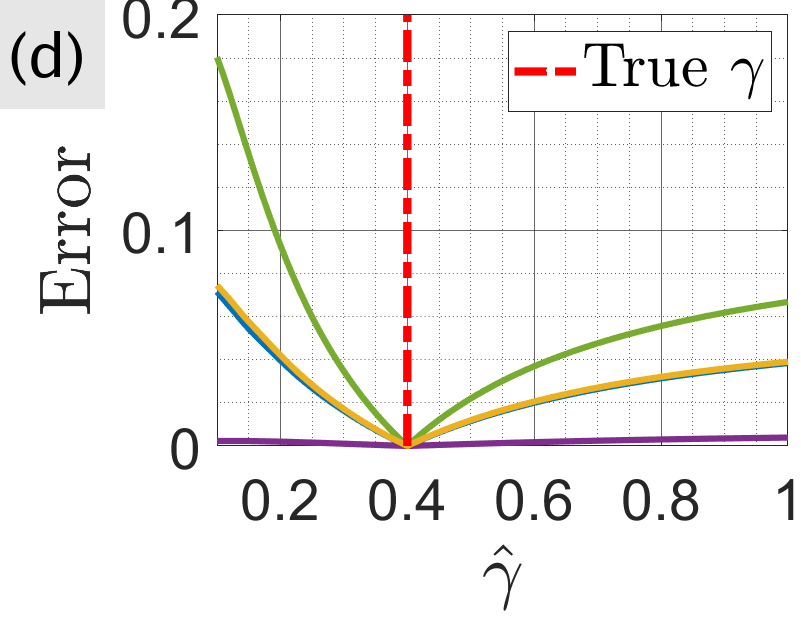}}\hfill\\
		\end{center}
	\end{center}
	\caption{For the no-noise case, the CRF can be estimated from any non-uniform patch. For the patches shown in (a), the linearisation error profiles without and with the normalization operation are shown in (c) and (d). We can observe that all the profiles show the correct global minima, and with the normalization operation, we can remove the local minima observed in (c) that occurs due to the scale problem.}
	\label{figure_synth_ill_1_cpall}
\end{figure*}

\subsection{Non-Linearity of Pixel Distributions}\label{sec_nonlinearity_dists}
From the input image $\mathbf{I}$, 	we can obtain a set of non-uniform patches. 
Let $\mathrm{\Omega}$ represent that set of the patches, and $|\mathrm{\Omega}|$ is  the number of the patches in the set. A patch drawn from this set is represented by $\mathbf{I}^{p}\in\mathrm{\Omega}$, where $p$ is the index of the drawn patch, and $p\in[1,|\mathrm{\Omega}|]$. 
Moreover, let the size of the patch $\mathbf{I}^p$ be $s \times s$; then,  $\mathbf{I}^{p,k}$ is a set of pixels in the patch (which we call a pixel distribution) where $k$ is the index of this set of pixels ($k\in[1, s]$).
Fig.~\ref{figure_edge_dists_img} shows an example of an input image  $\mathbf{I}$. 
Fig.~\ref{figure_edge_dists_patches} shows an example of some patches, $\mathbf{I}^p$, selected from  $\mathbf{I}$. 
Fig.~\ref{figure_edge_dists_patches} also shows the patches with a line in each patch, where the line represents the locations of pixels that form  a pixel distribution, $\mathbf{I}^{p,k}$.

Let $\mathbf{E}^{p,k}$ be the corresponding pixel distribution in the camera irradiance, and 
the three colour channels of $\mathbf{E}^{p,k}$ be represented by $\mathbf{E}^{p,k}_\text{r}$, $\mathbf{E}^{p,k}_\text{g}$ and $\mathbf{E}^{p,k}_\text{b}$.
When we transform $\mathbf{E}^{p,k}$ into the RGB space, it will form a straight line; since $\mathbf{E}^{p,k}_\text{r}$, $\mathbf{E}^{p,k}_\text{g}$ and $\mathbf{E}^{p,k}_\text{b}$ are linear to each other. Hence, we can express the linear correlation between two colour channels of the camera irradiance as:
\begin{align}
\label{eq_linear_Ep1}\mathbf{E}^{p,k}_\text{g} &\;=\; m^{p,k}_\text{gr}\:\mathbf{E}^{p,k}_\text{r} \;+\; b^{p,k}_\text{gr}, \\
\label{eq_linear_Ep2}\mathbf{E}^{p,k}_\text{b} &\;=\; m^{p,k}_\text{br}\:\mathbf{E}^{p,k}_\text{r} \;+\; b^{p,k}_\text{br},
\end{align}
where $(m^{p,k}_\text{gr}, b^{p.k}_\text{gr})$ and $(m^{p,k}_\text{br}, b^{p,k}_\text{br})$ are the parameters of the line equations.

Unlike $\mathbf{E}^{p,k}$ that forms a linear distribution, due to the non-linearity of the CRF, $\mathbf{I}^{p,k}$ forms a non-linear distribution in the RGB space, an observation introduced in~\cite{lin2004radiometric}. 
Fig.~\ref{figure_edge_dists_dists} shows an example of a few pixel distributions, $\{\mathbf{I}^{p,k}\}_{p=1}^4$,  plotted in the RGB space. We can observe the non-linearity in the distributions, which is due to the non-linear CRF shown in Fig.~\ref{figure_edge_dists_crf}.
The reason of the non-linearity of $\mathbf{I}^{p,k}$ in the RGB space can be explained as follows.
Since the three colour channels can have the same CRF and the linear operator $T$ can be a linear identity mapping, from Eq.~(\ref{eq_cam_model}), we can obtain: $\mathbf{E}^{p,k}_\text{r} = g(\mathbf{I}^{p,k}_\text{r})$, $\mathbf{E}^{p,k}_\text{g} = g(\mathbf{I}^{p,k}_\text{g})$ and $\mathbf{E}^{p,k}_\text{b} = g(\mathbf{I}^{p,k}_\text{b})$. Substituting these terms in Eqs.~(\ref{eq_linear_Ep1}) and (\ref{eq_linear_Ep2}), we can obtain the following non-linear equations:
\begin{align}
\label{eq_nonlinear_Ip1}\mathbf{I}^{p,k}_\text{g} &\;=\; g^{-1}(m^{p,k}_\text{gr}\:g(\mathbf{I}^{p,k}_\text{r}) \;+\; b^{p,k}_\text{gr}), \\
\label{eq_nonlinear_Ip2}\mathbf{I}^{p,k}_\text{b} &\;=\; g^{-1}(m^{p,k}_\text{br}\:g(\mathbf{I}^{p,k}_\text{r}) \;+\; b^{p,k}_\text{br}).
\end{align}
\bigbreak
\noindent{\textbf{No-Noise Case}}
Given any pixel distribution $\mathbf{I}^{p,k}$ from a non-uniform patch $\mathbf{I}^p$, where all the pixels are free from any noise and thus follow Eqs.~(\ref{eq_nonlinear_Ip1}) and  (\ref{eq_nonlinear_Ip2}) strictly, we can estimate the CRF by optimizing a CRF model $\hat{g}$ that can linearise the distribution with minimum error. The subsequent paragraph discusses this idea in more detail.

Fig.~\ref{figure_synth_ill_1_cp} shows four non-uniform patches taken from a synthetic image with no noise, which is generated using a simple CRF of a gamma function: $\mathbf{I} = f(\mathbf{E}) = \mathbf{E}^\gamma$ with $\gamma=0.4$.  Thus, for this case, the CRF model $\hat{g}$,  is the gamma model, where we need to find a gamma value $\hat{\gamma}$ to estimate the CRF. 
One of the simplest ways to find $\hat{\gamma}$ is to try all possible values of $\hat{\gamma}$ in a certain range; and for each of them, we generate the linearised distribution represented by $(\mathbf{I}^{p,k})^{1/\hat{\gamma}}$. Then, we can do a line fitting, and compute the error of each point from the line. This is basically the linearisation or line fitting error for $(\mathbf{I}^{p,k})^{1/\hat{\gamma}}$. 
Fig.~\ref{figure_synth_ill_1_cpprofs} shows the linearisation error profiles for the four pixel distributions taken from the four patches shown in Fig.~\ref{figure_synth_ill_1_cp}, respectively. 
The error profile for a pixel distribution is obtained by taking $\hat{\gamma}$ values in the range of $[0, 1]$ with increments of 0.02, and then computing the linearisation error for the distribution for all the $\hat{\gamma}$ values. 
From the results in Fig.~\ref{figure_synth_ill_1_cpprofs}, we can observe that all the error profiles show the correct global minima at $\hat{\gamma}=0.4$, with zero linearisation error. Note that, in this discussion, for the sake of clarity, we use only one pixel distribution, $\mathbf{I}^{p,k}$, for every patch, $\mathbf{I}^{p}$. In our actual algorithm, we use all the pixel distributions, $\{\mathbf{I}^{p,k}\}_{k=1}^s$, for every patch, $\mathbf{I}^{p}$, in the set $\mathrm{\Omega}$ (see Sec.~\ref{sec_consis_for_reliability}).

\begin{figure*}[t!]
	\captionsetup[subfloat]{labelformat=empty}
	\captionsetup[subfloat]{farskip=1pt}
	\begin{center}
		\begin{center}
			\subfloat[Patches]{\label{figure_synth_ill_1_np}\includegraphics[width=0.21\textwidth]{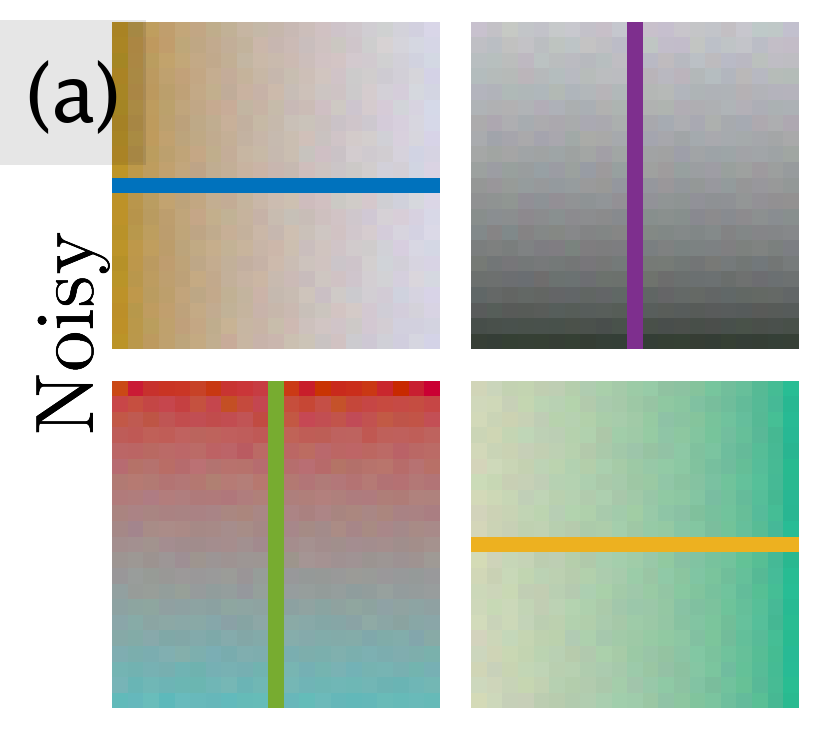}}\hfill
			\subfloat[Pixel Distributions]{\label{figure_synth_ill_1_npdists}\includegraphics[width=0.24\textwidth]{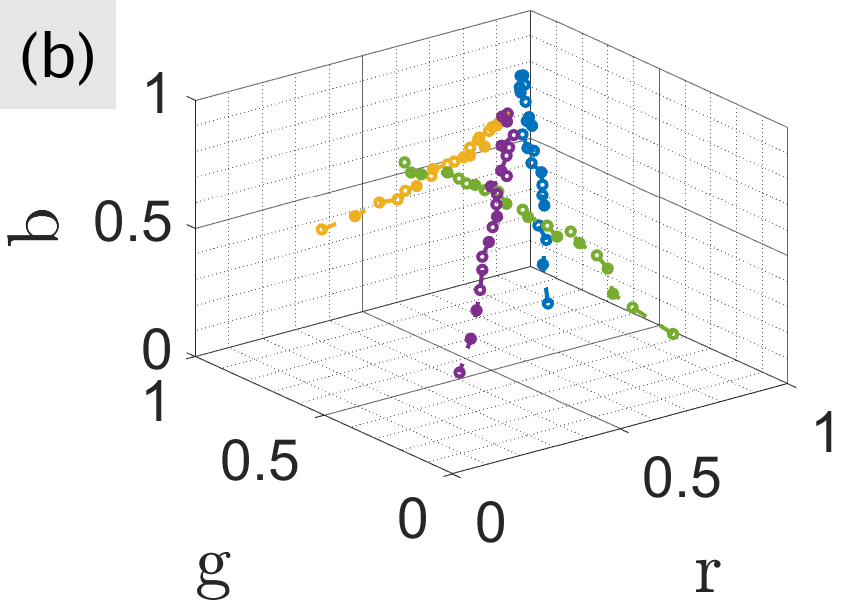}}\hfill
			\subfloat[Profiles (without $N$)]{\label{figure_synth_ill_1_npprofs}\includegraphics[width=0.24\textwidth]{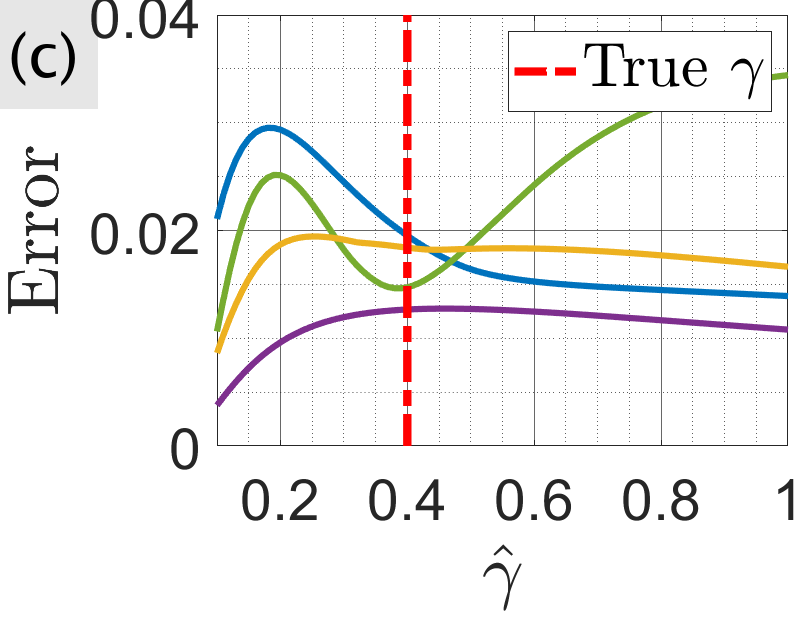}}\hfill
			\subfloat[Profiles (with $N$)]{\label{figure_synth_ill_1_npprofsN}\includegraphics[width=0.24\textwidth]{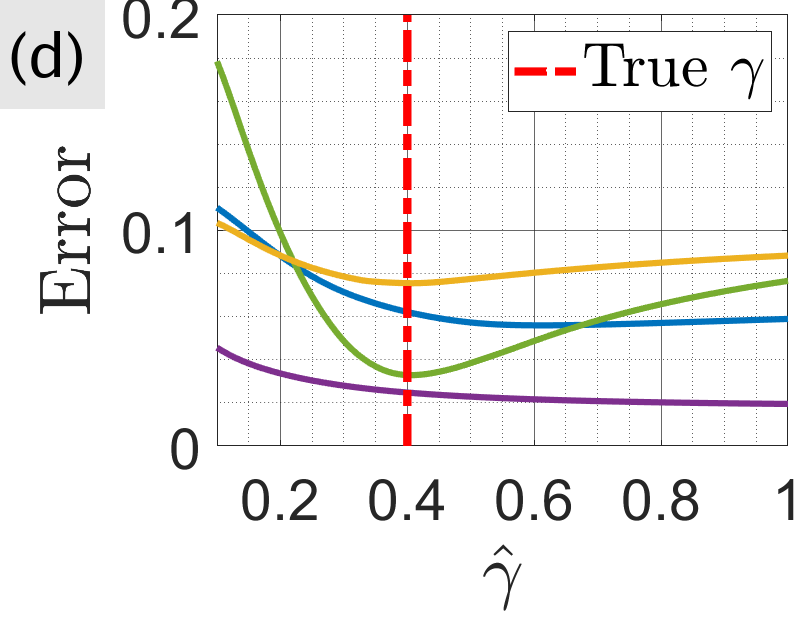}}\hfill\\
		\end{center}
	\end{center}
	\caption{For the noisy case, the linearisation error gets affected by both noise and non-linearity. For the noisy patches in (a), the profiles without the normalization operation shown in (c) show an incorrect global minima. As shown in (d), the normalization operation can alleviate this problem, but depending upon the noise and non-linearity, some distributions (like the purple and blue) can still provide an inaccurate result. }
	\label{figure_synth_ill_1_npall}
\end{figure*}

\bigbreak
\noindent{\textbf{Scale Problem and the Normalization Operation}}
From Fig.~\ref{figure_synth_ill_1_cpprofs}, we can also observe that all the error profiles  show local minima at $\hat{\gamma}=0.1$. The reason why this occurs is because the line fitting error is affected by the scale of $(\mathbf{I}^{p,k})^{1/\hat{\gamma}}$. As $\hat{\gamma}$ becomes small, the scale of $(\mathbf{I}^{p,k})^{1/\hat{\gamma}}$ also reduces; which in turn reduces the line fitting error. While this is not a significant problem for synthetic images with no noise, for real images where noise is inevitable, these local minima can possibly become global minima (see Fig.~\ref{figure_synth_ill_1_npprofs}). This is one of the reasons that some existing methods \cite{lin2004radiometric,lin2005determining} are erroneous.

To address this problem, we propose to normalise $(\mathbf{I}^{p,k})^{1/\hat{\gamma}}$ in each colour channel before computing the line fitting error. Namely, if $N$ represents the normalization operation, we normalise $(\mathbf{I}^{p,k})^{1/\hat{\gamma}}$ such that $\min(N((\mathbf{I}_\text{c}^{p,k})^{1/\hat{\gamma}}))$=0 and $\max(N((\mathbf{I}_\text{c}^{p,k})^{1/\hat{\gamma}}))$=1 for $\text{c}\in\{\text{r},\text{g},\text{b}\}$. The normalization operation makes the line fitting error independent of the scale variations caused by $\hat{\gamma}$ (since it is now the same scale for all values of $\hat{\gamma}$).  Fig.~\ref{figure_synth_ill_1_cpprofsN} shows  the profiles after the normalization operation, which now clearly indicate the correct global minima. 

\subsection{Effect of Noise on CRF Estimation}\label{sec_noiseaffect}
For the input image that has no noise, we can recover the CRF from any non-uniform patch, $\mathbf{I}^{p}$. 
However, this is not true when the image has noise. 
To illustrate this, we generate noisy patches (by adding random noise to the noiseless patches in Fig.~\ref{figure_synth_ill_1_cp}). 
Fig.~\ref{figure_synth_ill_1_np} shows the generated noisy patches, and Fig.~\ref{figure_synth_ill_1_npdists} shows the pixel distributions from the noisy patches plotted in the RGB space. 
Following the same procedure in the no-noise case, we compute the linearisation error profiles for the noisy pixel distributions. 
Fig.~\ref{figure_synth_ill_1_npprofs} and Fig.~\ref{figure_synth_ill_1_npprofsN} show the error profiles obtained without and with the normalization operation, respectively. 

As shown in Fig.~\ref{figure_synth_ill_1_npprofs}, we can observe that for the noisy patches, most profiles do not show the correct  global minima. This is because under the presence of noise, both noise and non-linearity affect the linearisation error. Moreover, we can observe that all the profiles show the global minima at $\hat{\gamma}=0.1$, which occurs because of the scale problem. 
Fig.~\ref{figure_synth_ill_1_npprofsN} shows the profiles after the normalisation operation that removes the scale problem. As a result, some profiles show the correct global minima, and some do not. This depends on the level of noise and the degree of non-linearity in the pixel distribution in the RGB space.

Eqs.~(\ref{eq_nonlinear_Ip1}) and (\ref{eq_nonlinear_Ip2}) show that the curviness of a patch (or its distribution $\mathbf{I}^{p,k}$) depends on its colour combination, and thus not all patches carry the CRF information equally. 
For instance, if the irradiance is achromatic or nearly achromatic, i.e.  $\mathbf{E}^{p,k}_\text{r} \approx \mathbf{E}^{p,k}_\text{g} \approx \mathbf{E}^{p,k}_\text{b}$, implying $m^{p,k}_\text{gr} \approx m^{p,k}_\text{br} \approx 1$ and $b^{p,k}_\text{gr} \approx b^{p,k}_\text{br} \approx 0$, Eqs.~(\ref{eq_nonlinear_Ip1}) and (\ref{eq_nonlinear_Ip2}) reduce to $\mathbf{I}^{p,k}_\text{r} \approx \mathbf{I}^{p,k}_\text{g} \approx \mathbf{I}^{p,k}_\text{b}$. This means that the line becomes linear for achromatic patches (or achromatic pixel distributions), and they have no information about the CRF.
As can be observed in Fig.~\ref{figure_synth_ill_1_cpprofsN}, the purple coloured profile of the pixel distribution that belongs to a nearly achromatic patch has a considerably shallow minima basin, which means that the distribution is close to being linear. 
This is what we mean by the degrees of non-linearity of patches (or pixel distributions) are unequal.

Moreover, from Eqs.~(\ref{eq_nonlinear_Ip1}) and (\ref{eq_nonlinear_Ip2}), we can also observe that due to the presence of both $g$ and $g^{-1}$ in the same equation, there is a ``cancellation'' effect, which means that in general, the degree of non-linearity in the RGB space is reduced. In other words, the curviness of the pixel distribution in the RGB space is lessened, causing the global minimum to be less obvious. Hence, when the noise comes into the picture, the error profile can be more significantly influenced by the noise, rendering incorrect CRF estimation, which can be observed for the blue and purple coloured distributions in Fig.~\ref{figure_synth_ill_1_npprofsN}. 
Therefore, we need some reliability measure to know the reliability score (which can indicate the degree of genuine non-linearity) of every patch in our set, so that we can rely more on the patches with higher reliability scores to estimate the CRF robustly and accurately. Note that most of the existing methods, particularly \cite{lin2004radiometric,lin2005determining}, assume insignificant noise in the input image and also ignore the patches' varying degrees of non-linearity, and thus treat all the processed patches equally.

\subsection{Prediction Consistency for Reliability}\label{sec_consis_for_reliability} 
To estimate the reliability score of a patch, we propose to use the consistency of the CRF predictions of all the pixel distributions in the patch. In a patch $\mathbf{I}^p$ of resolution $s\times s$, we have $s$ horizontally or vertically scanned pixel distributions, which can provide $s$ estimations of the CRF. If the estimations are more consistent, the  reliability score will be higher.

By employing the GGCM model~\cite{ng2007using} as our CRF model, we can express:
\begin{eqnarray}
\mathbf{I} = f(\mathbf{E}) = (\mathbf{E})^{\gamma_1+\gamma_2\mathbf{E}+...+\gamma_c(\mathbf{E})^{c-1}},	
\end{eqnarray}		
where $c$ is the number of coefficients, and $\{\gamma_1, ..., \gamma_c\}$ are the coefficients of the model.	
Thus by using the estimated coefficients, $\{\hat{\gamma}_1,..\hat{\gamma}_c\}$, we can obtain the estimated CRF represented by: $\hat{g}(\mathrm{x})=(\mathrm{x})^{\frac{1}{\hat{\gamma}_1+\hat{\gamma}_2\mathrm{x}+...+\hat{\gamma}_c(\mathrm{x})^{c-1}}}\:, \forall \mathrm{x}\in\mathrm{x}$, where  
$\mathrm{x}$ is a set of 100 equidistant values in the range of $[0,1]$. 
Let $\hat{g}^{p,k}(\mathrm{x})$ be the estimated CRF from a pixel distribution $\mathbf{I}^{p,k}$ in a patch. Then, we can compute the consistency between the $s$ estimations of the CRF, $\{\hat{g}^{p,k}(\mathrm{x})\}_{k=1}^s$, using:
\begin{equation}\label{eq_patch_consis}
\hat{\sigma}^{p} = \frac{1}{\vert \mathrm{x}\vert}\sum_{\mathrm{x}\in\mathrm{x}}\Biggl(\frac{1}{\vert s\vert}\sum_{k=1}^{s}\biggl(\hat{g}^{p,k}(\mathrm{x})-\frac{1}{\vert s\vert}\sum_{k=1}^{s}\left(\hat{g}^{p,k}(\mathrm{x})\right)\biggr)^2\Biggr).
\end{equation}
The reliability score of the patch, $\hat{\alpha}^p$, where:  $\hat{\alpha}^p=\exp^{\frac{-\hat{\sigma}^{p}}{0.05}}$. 
This score will be used in our gradual refinement scheme to weigh the CRF estimate of the corresponding patch (Sec.~\ref{sec_gradual_refinement}).

\begin{figure*}[t!]
	\captionsetup[subfloat]{labelformat=empty}
	\captionsetup[subfloat]{farskip=1pt}
	\begin{center}
		\subfloat{\includegraphics[width=0.33\textwidth]{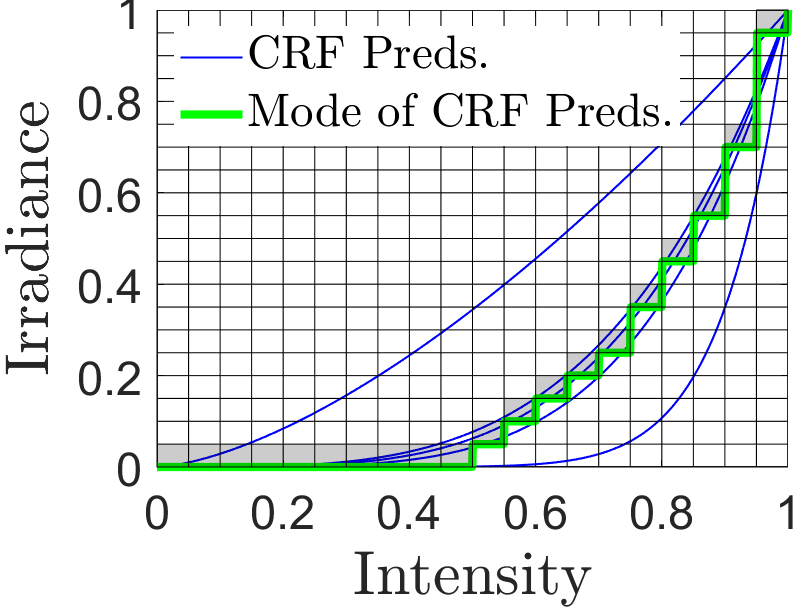}}\hfill
		\subfloat{\includegraphics[width=0.33\textwidth]{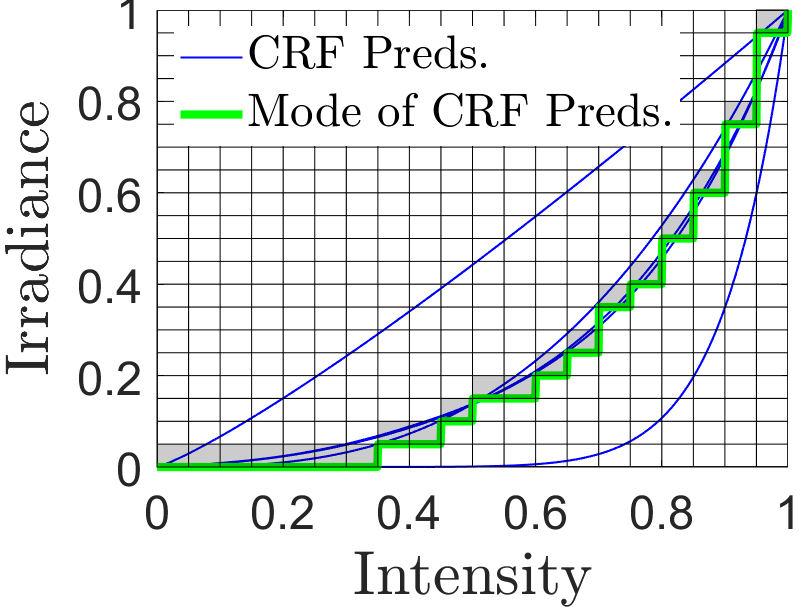}}\hfill
		\subfloat{\includegraphics[width=0.33\textwidth]{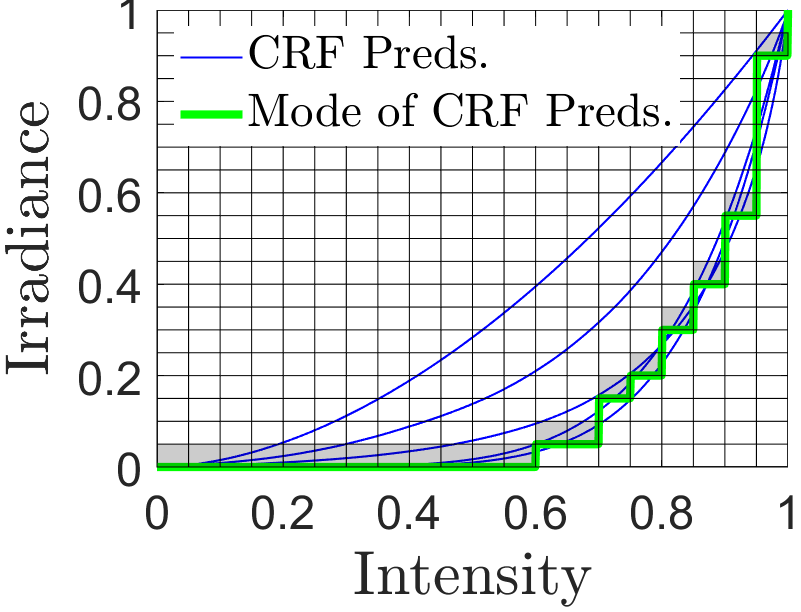}}\hfill\\
	\end{center}
	\caption{Examples showing the voting mechanism used for obtaining the mode of the CRF predictions. The mode CRF being discretised looks like a staircase function.}\label{figure_mode_crf}
\end{figure*}

To compute the CRF estimate of a single patch, $\hat{g}^p(\mathrm{x})$, instead of using  the mean of the $s$ predictions, we use the mode of the $s$ predictions. Since, from our investigation, in the present of noise, the mean is more influenced by the variations in the predictions than the mode. 
Here are the details. For every pixel distribution, $\mathbf{I}^{p,k}$ in a patch $\mathbf{I}^{p}$, we can obtain a CRF estimated curve, $\hat{g}^{p,k}(\mathrm{x})$. If we have $s$ pixel distributions in the patch, then we have $s$ CRF estimated curves, $\{\hat{g}^{p,k}(\mathrm{x})\}_{k=1}^s$.
We discretise the CRF space (as shown in the examples in Fig.~\ref{figure_mode_crf}) by creating a grid in the space. The grid is a  $\mathrm{\Delta} \times \mathrm{\Delta} $ grid. Meaning, 
there are $\mathrm{\Delta}$ columns and $\mathrm{\Delta}$ rows (where in our implementation $\mathrm{\Delta}=20$, hence our grid has $20\times20$ cells).  Subsequently, we count how many $\{\hat{g}^{p,k}(\mathrm{x})\}_{k=1}^s$ that fall into each of the cells. For each column in the grid, we choose the row that has the highest count. If we do this for all columns, we can have our discretised CRF estimate.
Mathematically, we express this discretised CRF estimate as:
\begin{equation}\label{eq_patch_pred}
\hat{g}^p(\mathrm{x}) = H\left(\{\hat{g}^{p,1}(\mathrm{x}), \hat{g}^{p,2}(\mathrm{x}), ..., \hat{g}^{p,s}(\mathrm{x})\}, \mathrm{\Delta}\right),
\end{equation}
where $H$ represents a function that chooses the best CRF for every column or intensity value in the discretised CRF space. In this space, the more correct cells contain higher values (see the examples shown in Fig.~\ref{figure_mode_crf}).

\subsection{Gradual Refinement}\label{sec_gradual_refinement} 

As we discussed, a CRF model with more coefficients can be more accurate in representing the CRF. However, under the presence of noise, a model with more coefficients is also more prone to overfit to noise. 
The existing methods (\cite{lin2004radiometric,lin2005determining,ng2007using}) use a one-attempt optimization approach to directly optimise a complex model (model with a high number of coefficients) which under the presence of noise can cause instability.
To address this problem, we propose to initialise the CRF estimation using the one-coefficient model (i.e., a model with only one coefficient). The initial result is more robust to noise but less accurate. We then increase the coefficients of the model to estimate more refined CRF result, which we constrain to remain near the previous CRF result. This is the core idea of our gradual refinement scheme. Also, note that while limiting the values of higher-order coefficients can also act as a regularizer, it can suppress the CRF representation capability, since some CRFs can require high values of the coefficients. Our gradual refinement, however, does not have this problem.

At stage $t$ of the refinement process, where $t\in[1,T]$, for a patch $\mathbf{I}^p$, we obtain the CRF prediction $\hat{g}^{p,k}_t(\mathrm{x})$ for its pixel distribution $\mathbf{I}^{p,k}$ by optimising the following objective function:
\begin{equation}\label{eq_stage_dist_pred}
\thickmuskip=0.2mu
\thinmuskip=0.2mu
\begin{split}
\hat{g}^{p,k}_t(\mathrm{x})\equiv \underset{\{\hat{\gamma}_1^{p,k},.. \hat{\gamma}_c^{p,k}\}}{\arg\!\min} &\biggl(\mathcal{L}\bigl[N\bigl((\mathbf{I}^{p,k})^{\frac{1}{\hat{\gamma}_1^{p,k}+...+\hat{\gamma}_c^{p,k}(\mathbf{I}^{p,k})^{c-1}}}\bigr)\bigr] \:\:+\\[-1ex] &\lambda\bigl((\mathrm{x})^\frac{1}{\hat{\gamma}_1^{p,k}+...+\hat{\gamma}_c^{p,k}(\mathrm{x})^{c-1}} - \hat{g}_{t-1}(\mathrm{x})\bigr)^2\biggr),
\end{split}
\end{equation}
where $c$ is the number of coefficients at stage $t$, $N$ is the normalization operation, $\mathcal{L}$ is a function that computes the line fitting error, and $\lambda$ is a parameter that controls the closeness of the CRF prediction $\hat{g}^{p,k}_t(\mathrm{x})$ to the CRF estimated in the previous stage $\hat{g}_{t-1}(\mathrm{x})$. We keep $\lambda$=0 at $t$=1. Having obtained the predictions from the patch's pixel distributions, we compute the reliability score of the patch, $\hat{\alpha}^p_t$, and the CRF estimate from the patch, $\hat{g}^p_t(\mathrm{x})$, (as described in Sec.~\ref{sec_consis_for_reliability}). 

The CRF estimate at stage $t$ from all the patches, $\hat{g}_t(\mathrm{x})$, is then computed by considering the reliability scores of the patches:
\begin{equation}\label{eq_stage_pred}
\thickmuskip=3mu
\thinmuskip=3mu
\hat{g}_t(\mathrm{x}) = H\left(\{\hat{\alpha}^1_t\hat{g}^{1}_t(\mathrm{x}), \hat{\alpha}^2_t\hat{g}^{2}_t(\mathrm{x}), ..., \hat{\alpha}^{|\mathrm{\Omega}|}_t\hat{g}^{|\mathrm{\Omega}|}_t(\mathrm{x})\}, \mathrm{\Delta}\right),
\end{equation}
where we use the same voting mechanism, except instead of counting the number of predictions for each grid cell, we sum the reliability scores of the predictions for each grid cell. For each column, the row that contains the largest sum is selected.

Practically, to improve the CRF estimation accuracy in the next stage and to keep the runtime of the method small, we remove the patches in $\mathrm{\Omega}$ whose reliability scores are lower than a certain threshold, $\tau_\text{re}$. 
After completion of stage $T$, i.e. at the end of the refinement process, the estimated CRF $\hat{g}_T(\mathrm{x})$ is a staircase curve. To generate a smooth CRF curve, we fit our $c$-coefficient CRF model on $\hat{g}_T(\mathrm{x})$. 
\begin{equation}\label{eq_final_pred}
\hat{g}(\mathrm{x})\equiv \underset{\{\hat{\gamma}_1,.. \hat{\gamma}_c\}}{\arg\!\min} \:\:\left((\mathrm{x})^{\frac{1}{\hat{\gamma}_1+\hat{\gamma}_2\mathrm{x}+...+\hat{\gamma}_c(\mathrm{x})^{c-1}}} - \hat{g}_{T}(\mathrm{x})\right)^2, 
\end{equation}
where $\hat{g}(\mathrm{x})$ is the final CRF result from our method. See Algorithm~\ref{algo_overall} for our entire CRF estimation process.

\begin{algorithm}[t!]
	\begin{algorithmic}[1]
		\State \textbf{Input:} Image $\mathbf{I}$. 
		\State Generate a set of non-uniform patches $\mathrm{\Omega}$ following the selection process in Sec.~\ref{sec_patch_selection}.
		\For {stage t where $t\in[1, T]$}
		\For {patch $\mathbf{I}^p$ of resolution $s\times s$ where $\mathbf{I}^p\in\mathrm{\Omega}$}
		\For {patch distribution $\mathbf{I}^{p,k}$ where $k\in[1, s]$}
		\State\parbox[t]{190pt}{%
 			Obtain the CRF estimate for the distribution, $\hat{g}^{p,k}_t(\mathrm{x})$, using Eq.~(\ref{eq_stage_dist_pred}).\strut}
		\EndFor
		\State\parbox[t]{205pt}{% 
			For the patch, obtain the CRF estimate, $\hat{g}^p_t(\mathrm{x})$, and reliability, $\hat{\alpha}^p_t$, using Eqs.~(\ref{eq_patch_pred}) and (\ref{eq_patch_consis}) respectively.\strut}
		\EndFor
		\State\parbox[t]{190pt}{%  
			Obtain the CRF estimate at stage $t$, $\hat{g}_t(\mathrm{x})$, using Eq.~(\ref{eq_stage_pred}).\strut}
		\State Update $\mathrm{\Omega}$. Remove the patches whose $\hat{\alpha}^p_t<\tau_\text{re}$.
		\EndFor
		\State Obtain the final CRF, $\hat{g}(\mathrm{x})$, from $\hat{g}_T(\mathrm{x})$ using Eq.~(\ref{eq_final_pred}).
		\State \textbf{Output:} CRF $\hat{g}(\mathrm{x})$.   
	\end{algorithmic}
	\caption{Single-Image Camera Response Function using Prediction Consistency and Gradual Refinement}
	\label{algo_overall}
\end{algorithm}

\subsection{Selection of Patches}\label{sec_patch_selection} 
We add a patch into $\mathrm{\Omega}$ if it meets the following criteria:
\begin{enumerate}[noitemsep,topsep=1pt]
	\item The patch has no under-saturated or over-saturated pixels, i.e.  for every pixel in the patch, its magnitude (mean of the three colour channel values) is above $\tau_\text{us}$ and below $\tau_\text{os}$, where $\tau_\text{us}$ and $\tau_\text{os}$ are the thresholds set for the under-saturated and over-saturated pixels, respectively.
	\item The patch is not uniform and contains a mixture of colours, i.e. the variance of all the three colour channel values in the patch is above $\tau_\text{un}$, where $\tau_\text{un}$ is the threshold set for the uniformity of a patch.
	\item  The patch's pixel distributions are not narrow and are well spread in the RGB space, i.e the variance of a colour channel values in the patch is above $\tau_\text{na}$, where $\tau_\text{na}$ is the threshold set for narrowness of the distributions. 
\end{enumerate}	
To select the pixel distributions in a patch, we select the distributions by scanning in either horizontal or vertical direction, depending upon which direction the distributions have more variance.

Note that: (1) unlike ~\cite{lin2004radiometric}, which uses solely edge patches, our patch selection method also includes patches with any mixture of colours  beyond edges. This gives us a more rich set of patches; (2) The selection criteria above are learned empirically, and can be adjusted depending on the conditions of the target images; (3) Our horizontal-vertical scanning technique is not a hard requirement, as  any other techniques can also be used. 
%The only reason we opt for the simple scanning technique is to keep our method efficient.

\begin{figure}[t!]
	\vspace{-0.15in}
	\captionsetup[subfloat]{labelformat=empty}
	\captionsetup[subfloat]{farskip=1pt}
	\begin{center}
		\subfloat{\includegraphics[width=1.0\columnwidth]{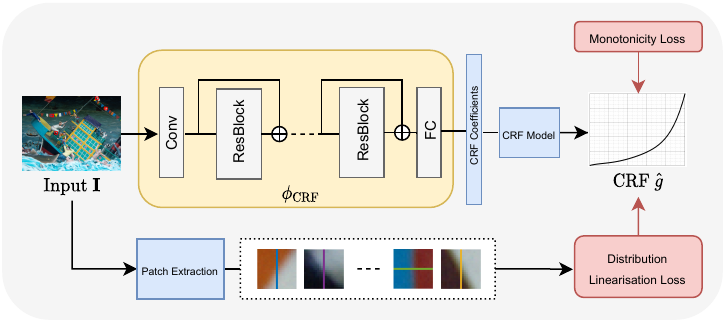}}
	\end{center}
	\caption{For the test real input image $\mathbf{I}$, our proposed network $\phi_\text{CRF}$ estimates the CRF $\hat{{g}}$ using test-time training based on unsupervised losses. The network is assumed to be already pretrained on synthetic labelled data using supervised losses.}\label{figure_dl_model}
\end{figure}

\subsection{Extension to Deep Learning}\label{sec_dl_extn} 
To improve the efficiency and practicality, we propose a deep learning extension. We propose a network to estimate the CRF that is trained in a semi-supervised manner. Specifically, we train our network in two stages: (1) Main training, where we use labelled synthetic images with supervised losses, and (2) Test-time training, where we use unsupervised losses to finetune the network on the test input image that improves the CRF result on the test image. We use ResNet-18~\cite{he2016deep} as the backbone of our network (see Fig.~\ref{figure_dl_model}).

We refer to our network by $\phi_\text{CRF}$. The set of labelled synthetic images and the set of their corresponding ground-truth inverse CRFs are represented by $\mathbb{J}$ and $\mathbbm{g}^{gt}$ respectively. Samples drawn from $\mathbb{J}$ and $\mathbbm{g}^{gt}$ are represented by $\mathbf{J}$ and ${g}^{gt}$ respectively. 

\smallbreak
\noindent\textbf{Main Training: Supervised}
In the main training stage, we use supervised losses on our labelled synthetic data to train our network. 
Specifically, given $\mathbf{J}$ as input and $g^{gt}$ as the ground-truth inverse CRF, we use the following loss:
\begin{equation}\label{eq_main_crf_mseloss}
\mathcal{L}_\text{mse} = \lVert\hat{g}(\mathrm{x}) - {g^{gt}(\mathrm{x})}\rVert_2, 
\end{equation} 
where $\hat{g}(\mathrm{x})$ is the predicted inverse CRF. The network predicts the coefficients, $\{\hat{\gamma}_1,..\hat{\gamma}_c\}$, which in turn provide the CRF estimate $\hat{g}(\mathrm{x})$ (i.e. $\{\hat{\gamma}_1,..\hat{\gamma}_c\} = \phi_\text{CRF}(\mathbf{J})$).
Since we also have the ground-truth linearised image, ${g^{gt}}(\mathbf{J})$, we use the linearisation loss to train $\phi_\text{CRF}$:
\begin{equation}\label{eq_main_crf_linloss}
\mathcal{L}_\text{lin} = \lVert\hat{{g}}(\mathbf{J}) - {{g}^{gt}(\mathbf{J})}\rVert_1. 
\end{equation} 

\smallbreak
\noindent\textbf{Test-time Training: Unsupervised}
Having trained the network $\phi_\text{CRF}$ on our labelled synthetic data, we finetune it on our test real input image $\mathbf{I}$ in test-time training using unsupervised losses (since $\mathbf{I}$ has no ground-truth CRF available). 

CRFs are known to be monotonically increasing~\cite{grossberg2004modeling}, hence we use the monotonicity loss~\cite{shi2010self}:
\begin{equation}\label{eq_testtime_monloss}
\mathcal{L}_\text{mon} = \sum_{t=0}^{1} H\left(-\frac{\partial \hat{{g}}(\mathrm{x})}{\partial \mathrm{x}}\right), 
\end{equation}
where we impose that the derivative of predicted inverse CRF $\hat{{g}}(\mathrm{x})$, $\partial \hat{{g}}(\mathrm{x})/\partial \mathrm{x}$, to be positive. $H(\cdot)$ is the Heaviside step function: $H(x)=1$ when $x\ge0$, and $0$ otherwise.

Following the same patch selection strategy mentioned in Sec.~\ref{sec_patch_selection}, we obtain a set of patches $\Omega$ from the input image $\mathbf{I}$; and for each patch $\mathbf{I}^p\in\Omega$, we obtain $s$ pixel distributions $\{\mathbf{I}^{p,k}\}_{k=1}^{s}$. 
The entire set of pixel distributions obtained from all the patches is represented by $\{\mathbf{I}^{11},...,\mathbf{I}^{1s},...,\mathbf{I}^{|\Omega|1},...,\mathbf{I}^{|\Omega|s}\}$. For each pixel distribution $\mathbf{I}^{p,k}$ where $p\in[1,|\Omega|]$ and $k\in[1,s]$, we first linearise it using the predicted inverse CRF $\hat{{g}}$, and then normalize it. 
The normalisation operation is important in order to avoid trivial solution (see Sec.~\ref{sec_nonlinearity_dists}). Representing the linearised and normalized distribution by $\mathbf{\tilde{I}}^{p,k}$ where $\mathbf{\tilde{I}}^{p,k} = N(\hat{g}(\mathbf{I}^{p,k}))$, and its minimum and maximum pixel values by $\mathbf{\tilde{I}}^{p,k}_{\text{min}}$ and $\mathbf{\tilde{I}}^{p,k}_{\text{max}}$ respectively, we define our linearisation loss for pixel distributions as:
\begin{gather}\label{eq_testtime_edgelinloss}
\mathcal{L}^{p,k}\!=\! \sum_{i=1}^{s}\!\bigg(\frac{\lvert(\mathbf{\tilde{I}}^{p,k}_{\text{min}}-\mathbf{\tilde{I}}^{p,k}_{\text{max}}) \times (\mathbf{\tilde{I}}^{p,k}_{\text{min}}-\mathbf{\tilde{I}}^{p,k,i})\rvert}{\lvert\mathbf{\tilde{I}}^{p,k}_{\text{min}}-\mathbf{\tilde{I}}^{p,k}_{\text{max}}\rvert}\bigg),\\
\mathcal{L}_\text{distlin} = \sum_{p=1}^{|\Omega|}\left(\sum_{k=1}^{s}\left(\mathcal{L}^{p,k}\right)\right),
\end{gather}
where $\mathbf{\tilde{I}}^{p,k,i}$ is the pixel value of the $i^\text{th}$ pixel on the pixel distribution $\mathbf{\tilde{I}}^{p,k}$. 

\begin{table}[!t]
	\centering
	\setlength\belowcaptionskip{-5pt}
	\renewcommand{\arraystretch}{1.2}
	\caption {Comparisons with the baselines on the daytime test dataset. The numbers represent RMSE. Bold font indicates lowest error} \label{table_daytime_results}		
	\begin{tabularx}{\columnwidth}{ c|Y|Y|Y|Y }
		\toprule
		Method & Mean & Std & Min & Max \\
		\midrule
		\phantom{zz}EdgeCRF~\cite{lin2004radiometric}\phantom{zz} &0.1561 &0.0402 &0.0424 &0.2522\\
		%\hline           
		CRFNet~\cite{li2017crf} &0.1609 &0.0538 &0.0459 &0.2788\\
		%\hline           
		GICRF~\cite{ng2007using} &0.0943 &0.0340 &0.0379 &0.2415\\
		%\hline
		\textbf{Our Method} &\textbf{0.0406} &\textbf{0.0308} &\textbf{0.0142} &\textbf{0.2077}\\
		\bottomrule
		\vspace{0.01in}
	\end{tabularx}
\end{table}

\begin{table}[!t]
	\centering
	\setlength\belowcaptionskip{-5pt}
	\renewcommand{\arraystretch}{1.2}
	\caption {Comparisons with baselines on the nighttime test dataset. The numbers represent RMSE. Bold font indicates lowest error} \label{table_nighttime_results}		
	\begin{tabularx}{\columnwidth}{ c|Y|Y|Y|Y }
		\toprule
		Method & Mean & Std & Min & Max \\
		\midrule
		%\hline
		\phantom{zz}EdgeCRF~\cite{lin2004radiometric}\phantom{zz} &0.1738  &0.0504 &0.0593 &0.2629\\
		%\hline           
		CRFNet~\cite{li2017crf} &0.2407  &0.1059 &0.0681 &0.4428\\
		%\hline           
		GICRF~\cite{ng2007using} &0.1149  &0.0533 &0.0669 &0.2819\\
		%\hline
		\textbf{Our Method} &\textbf{0.0521}  &\textbf{0.0350} &\textbf{0.0173} &\textbf{0.1996}\\
		\bottomrule           
	\end{tabularx}	
\end{table}

\begin{figure*}[t!]
	\captionsetup[subfloat]{labelformat=empty}
	\captionsetup[subfloat]{farskip=1pt}
	\setlength{\belowcaptionskip}{-12pt}
	\begin{center}
		\begin{center}
			\subfloat{\includegraphics[width=0.235\textwidth]{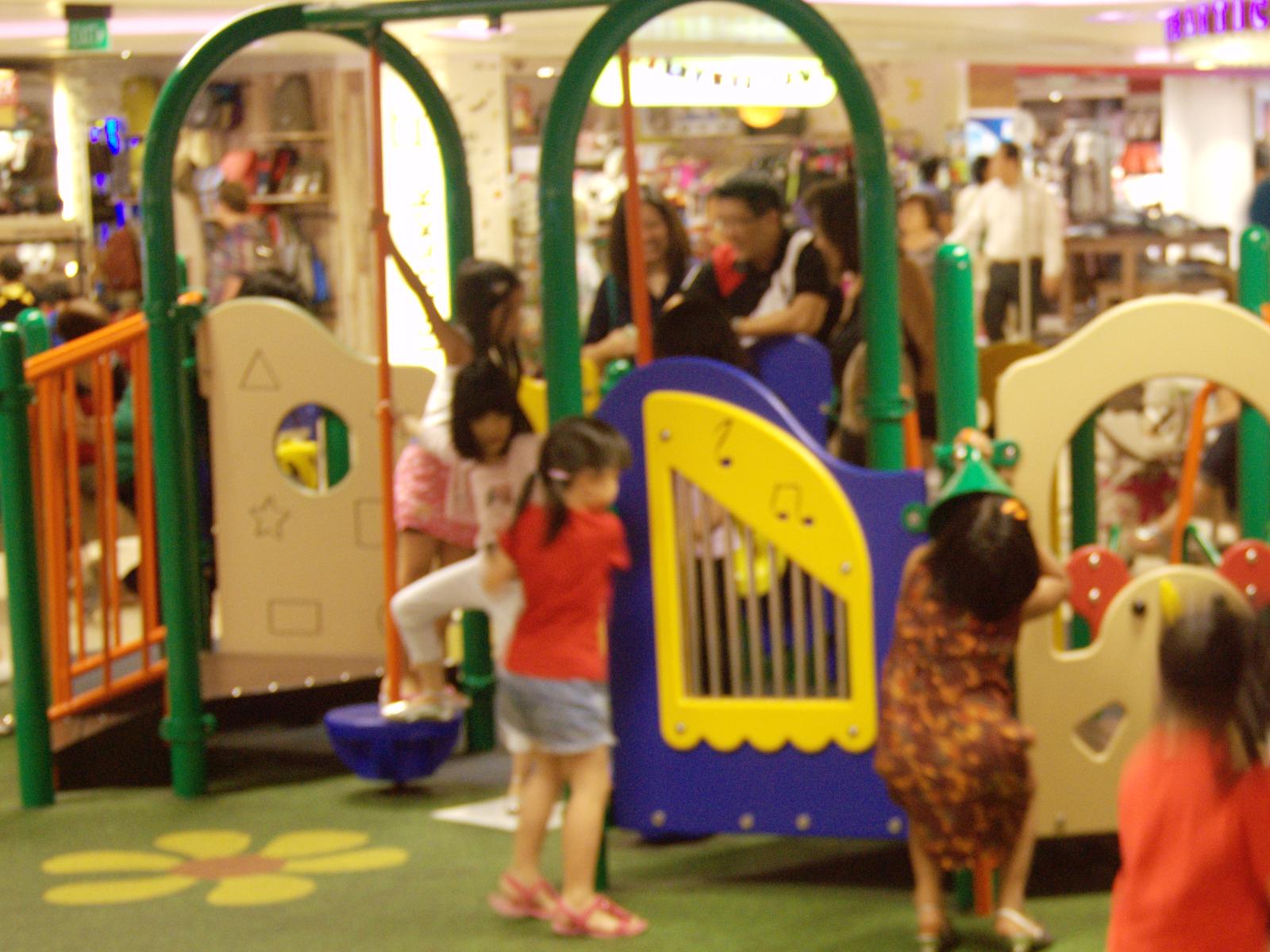}}
			\subfloat{\includegraphics[width=0.235\textwidth]{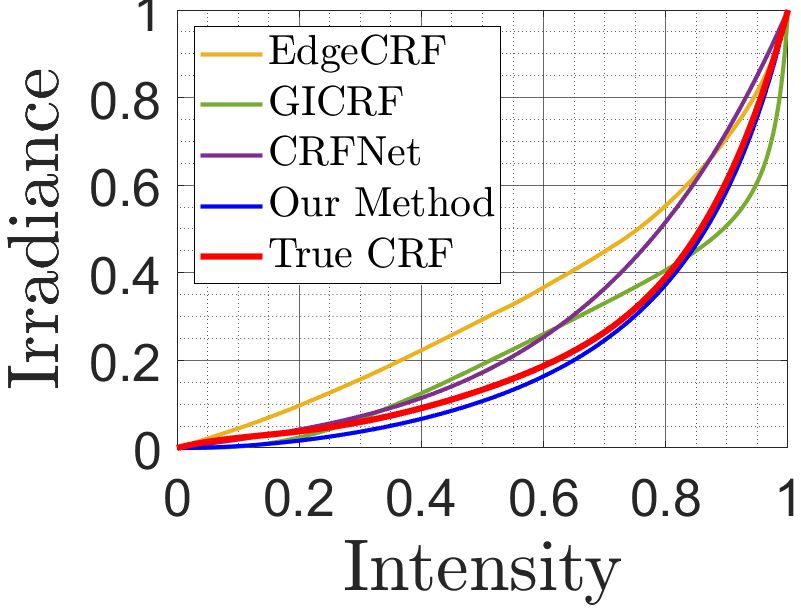}}\hfill
			\subfloat{\includegraphics[width=0.235\textwidth]{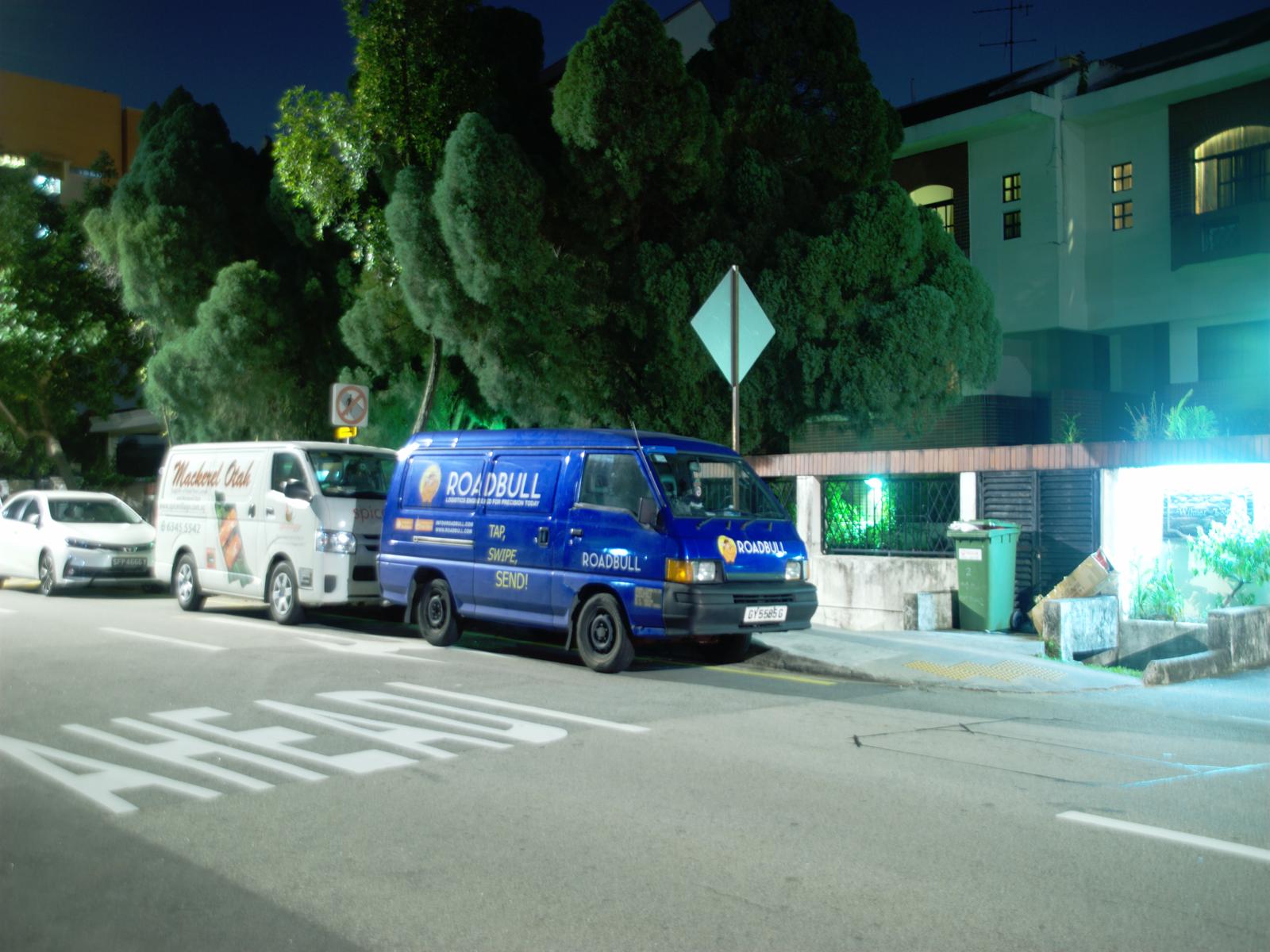}}
			\subfloat{\includegraphics[width=0.235\textwidth]{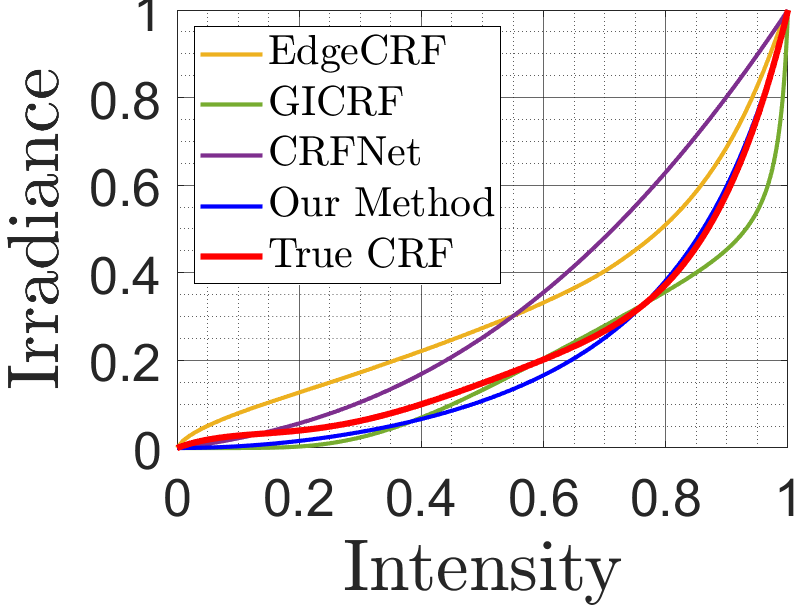}}\hfill\\
			
			\subfloat{\includegraphics[width=0.235\textwidth]{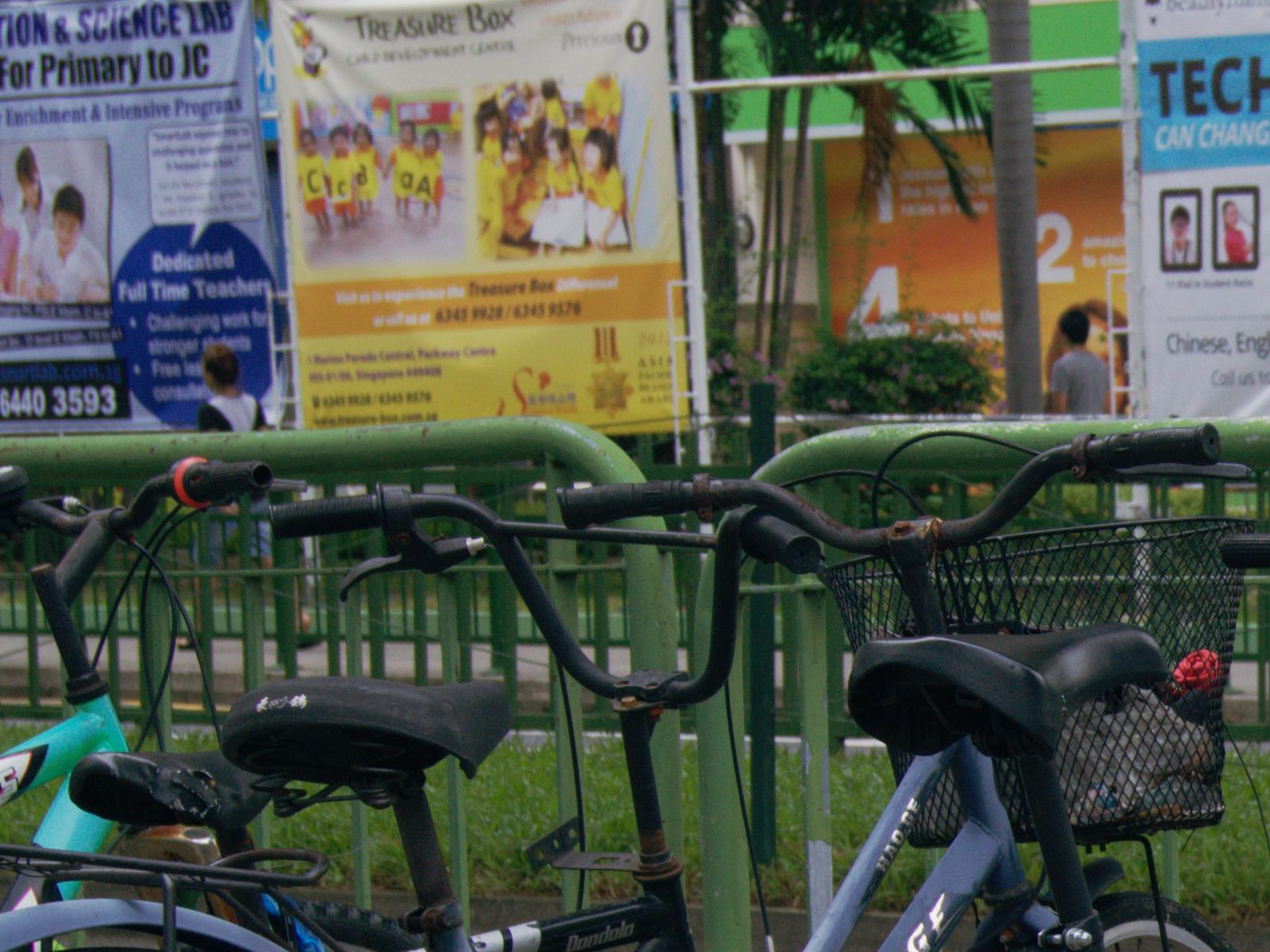}}
			\subfloat{\includegraphics[width=0.235\textwidth]{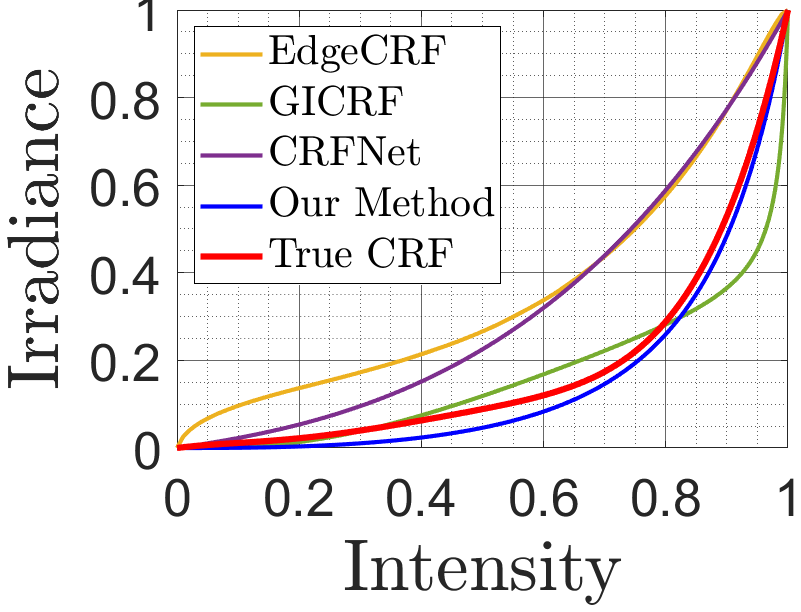}}\hfill
			\subfloat{\includegraphics[width=0.235\textwidth]{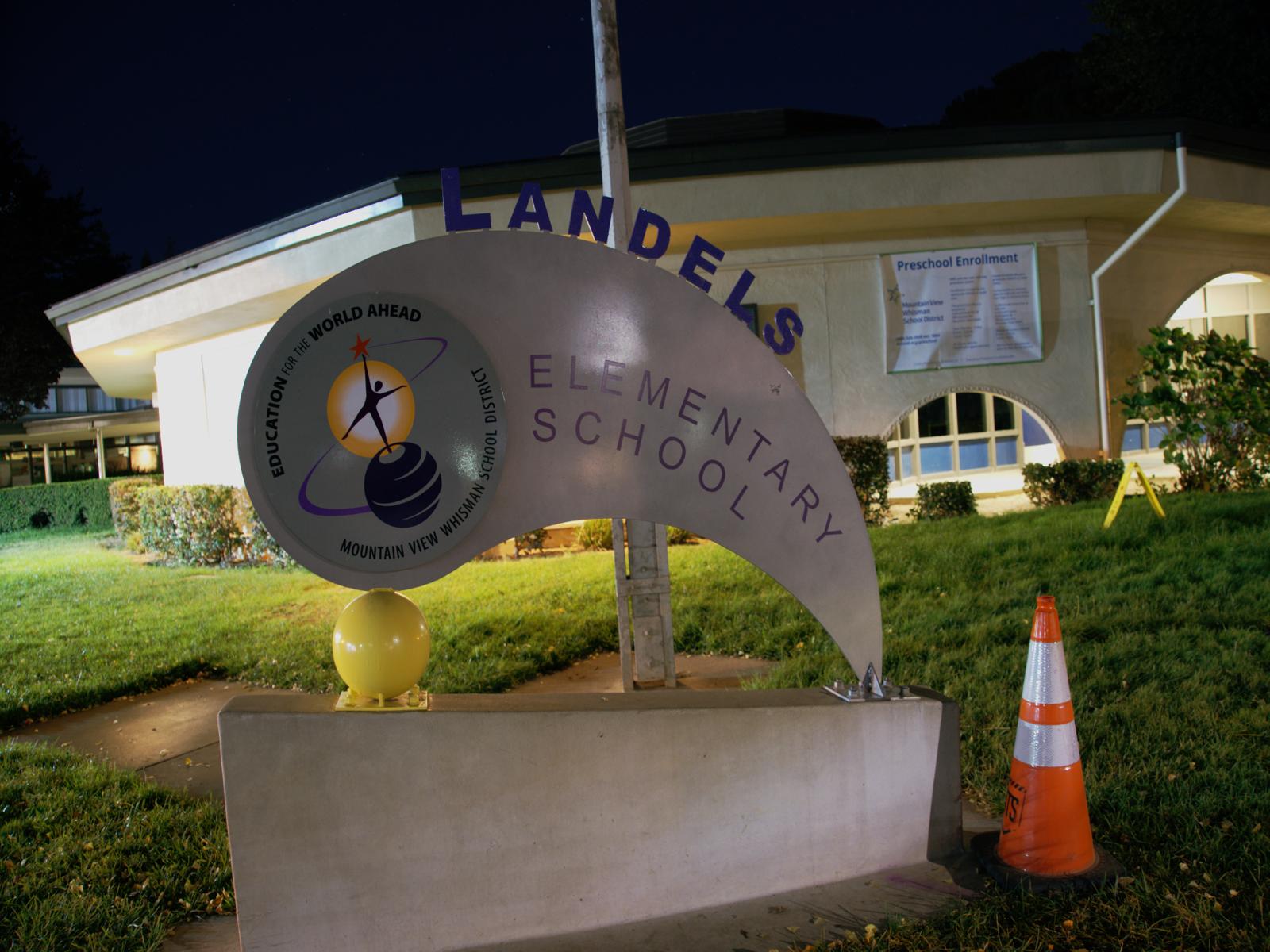}}
			\subfloat{\includegraphics[width=0.235\textwidth]{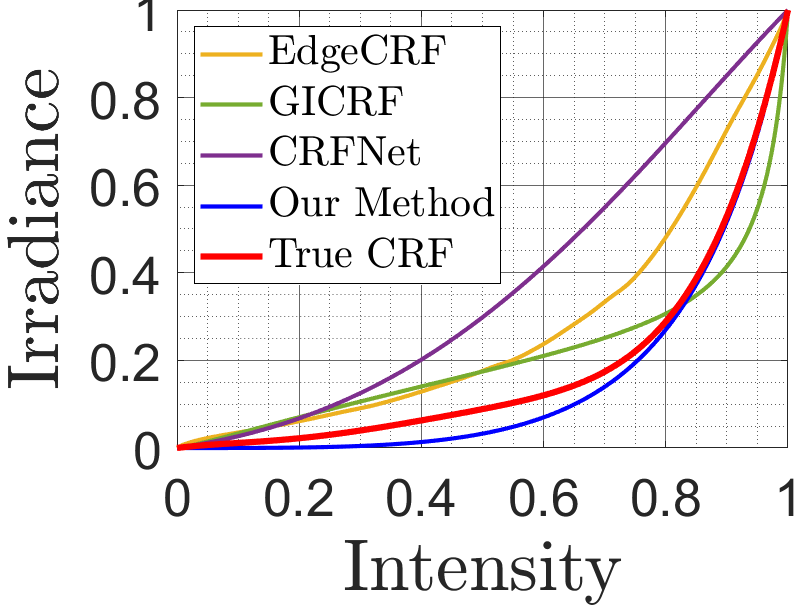}}\hfill\\
			
			\subfloat[Input Image]{\includegraphics[width=0.235\textwidth]{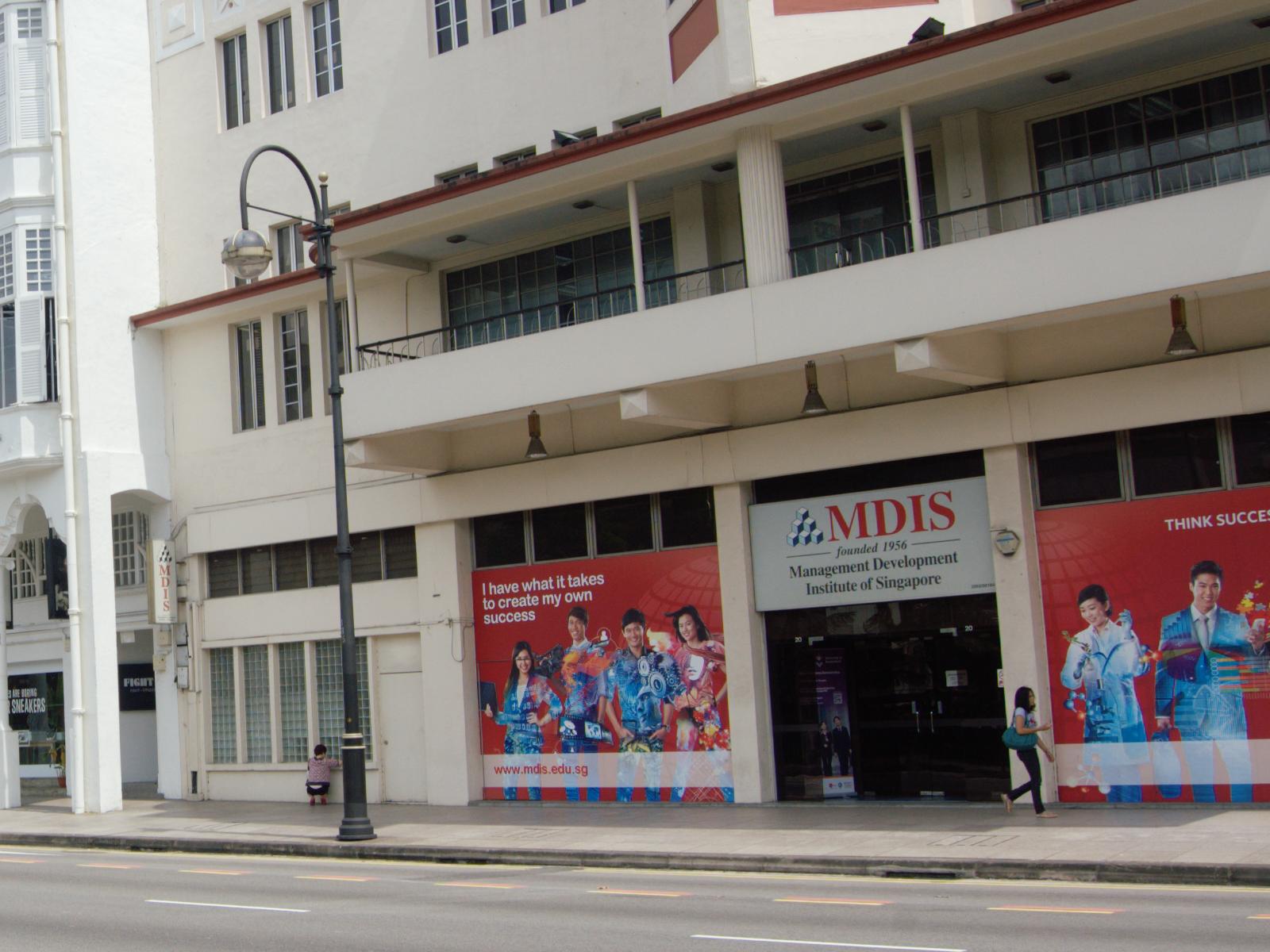}}
			\subfloat[Predicted CRF]{\includegraphics[width=0.235\textwidth]{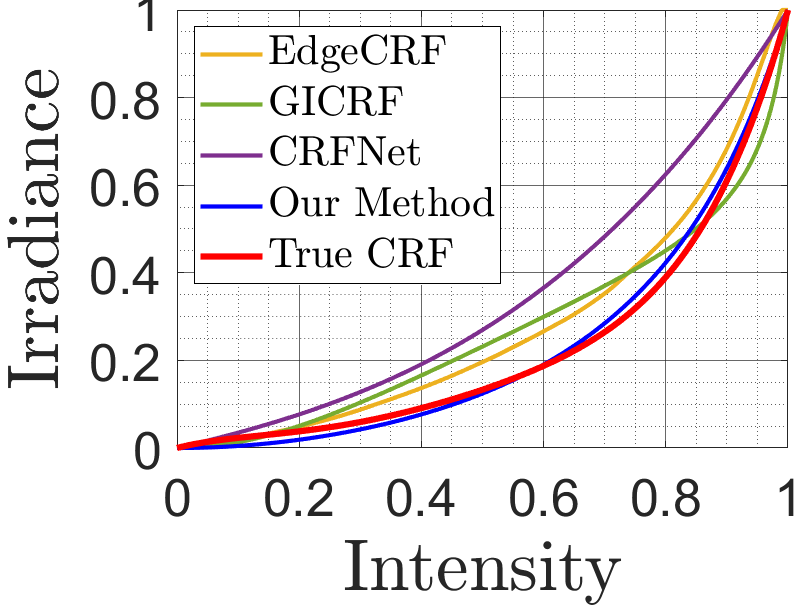}}\hfill			
			\subfloat[Input Image]{\includegraphics[width=0.235\textwidth]{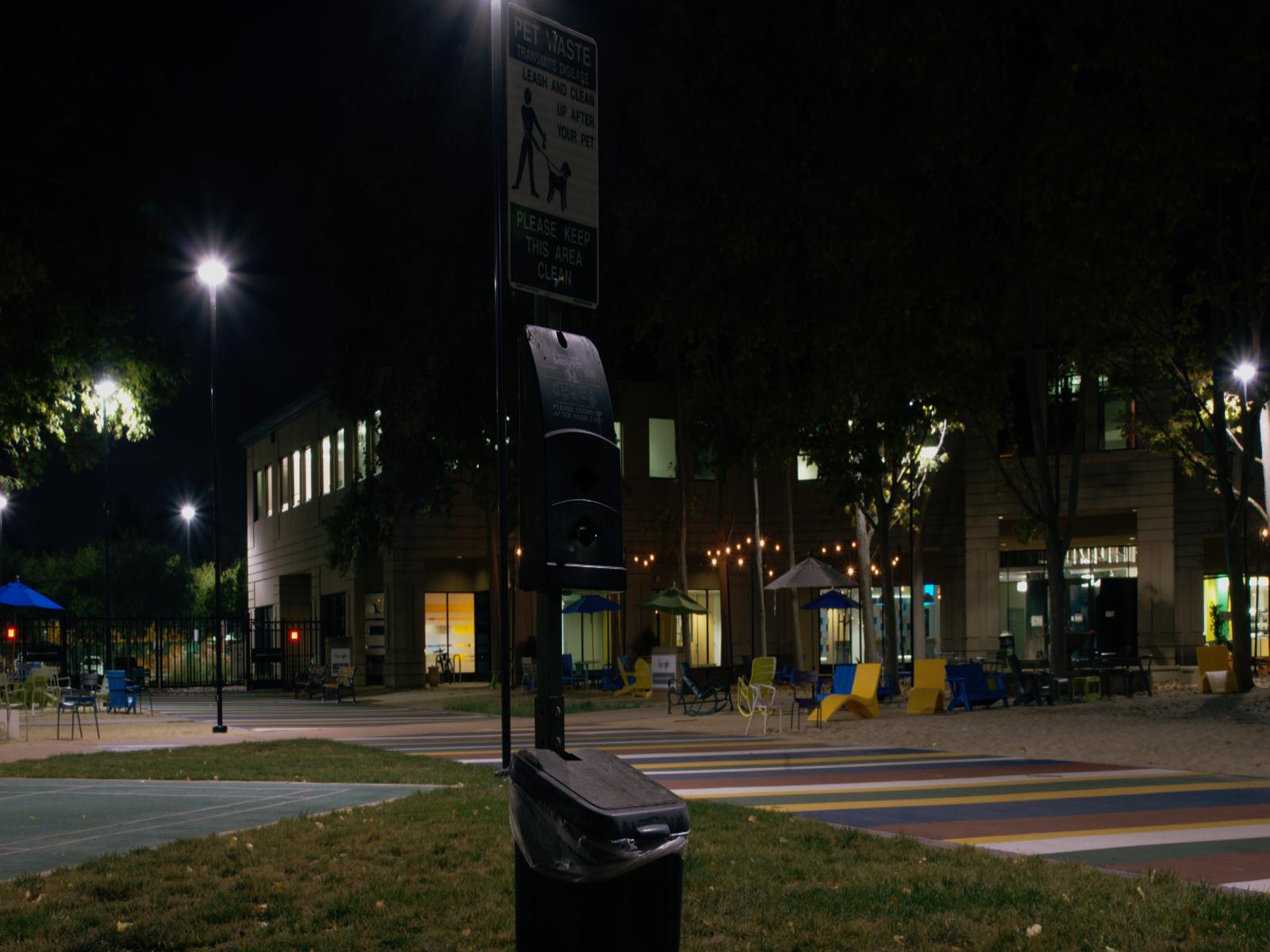}}
			\subfloat[Predicted CRF]{\includegraphics[width=0.235\textwidth]{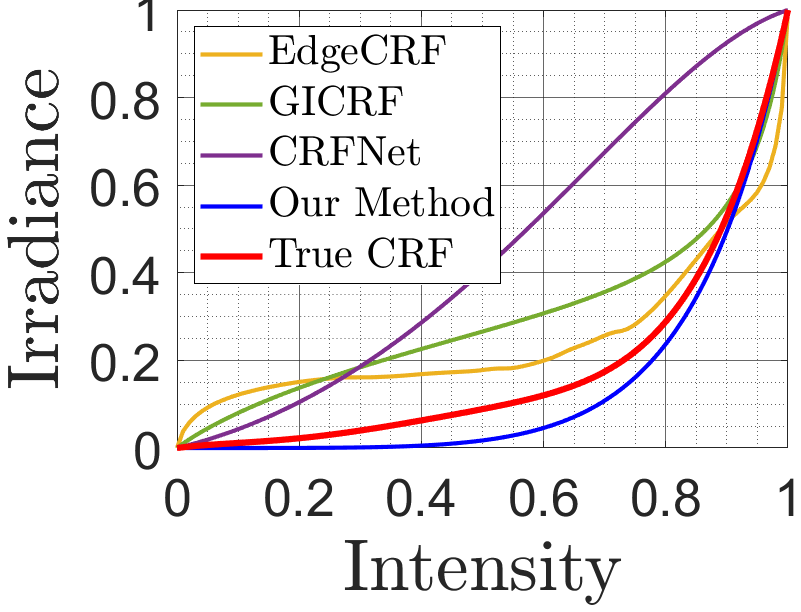}}\hfill\\
		\end{center}
	\end{center}
	\caption{Qualitative comparisons with the baseline methods. As we can observe, compared to the baseline methods, our results are closest to the true CRF, showing our method's robustness in both general daytime and nighttime conditions.}\label{figure_daynighttime_results}
\end{figure*}

\begin{figure*}[t!]
	\vspace{0.15in}
	\captionsetup[subfloat]{labelformat=empty}
	\captionsetup[subfloat]{farskip=1pt}
	%\captionsetup[subfloat]{position=top}
	\begin{center}
		\subfloat[{Input Image (PSNR: 11.44)}]{\includegraphics[width=0.247\textwidth]{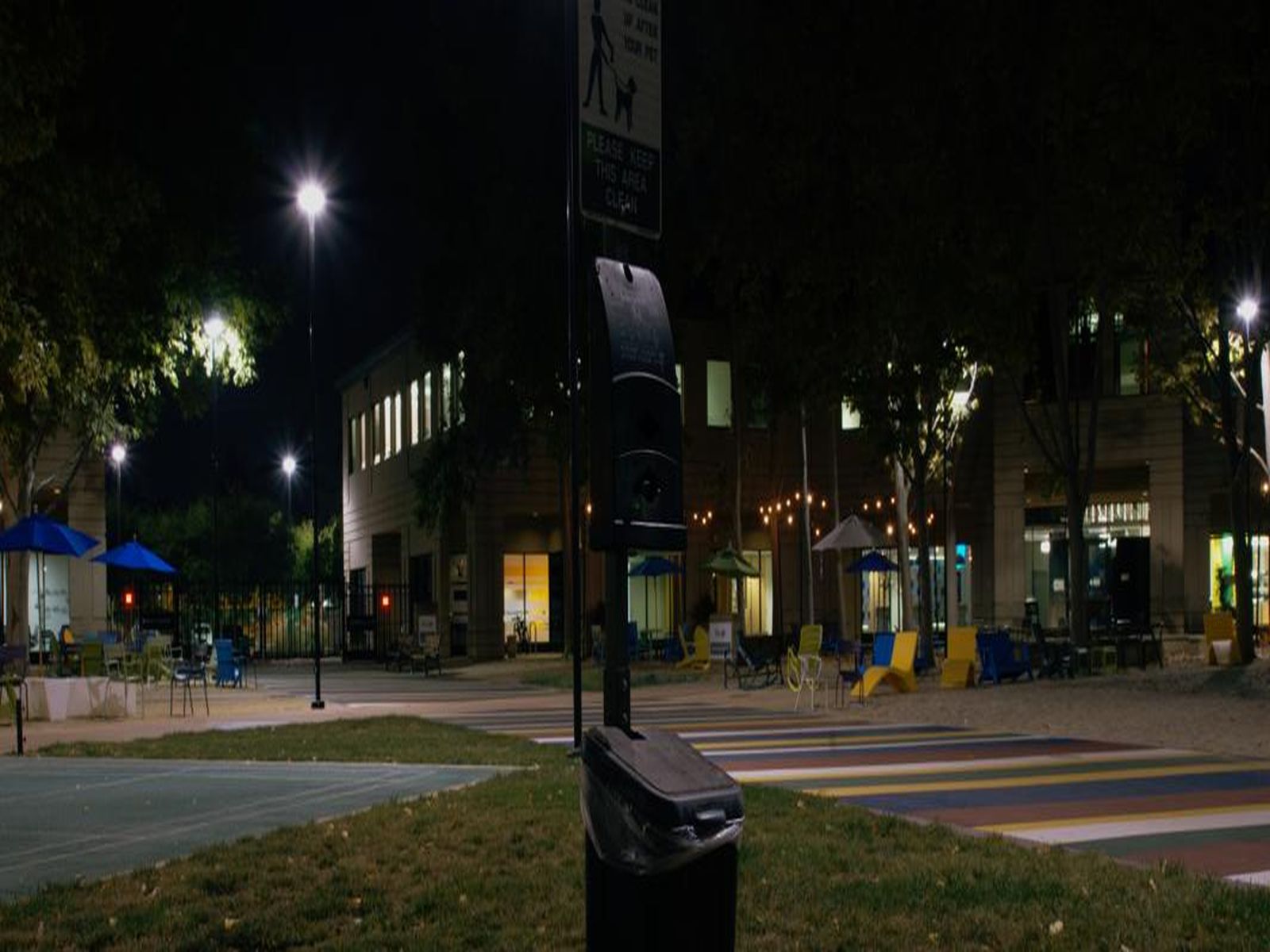}}\hfill
		\subfloat[LIME+BM3D (PSNR: 17.23)]{\includegraphics[width=0.247\textwidth]{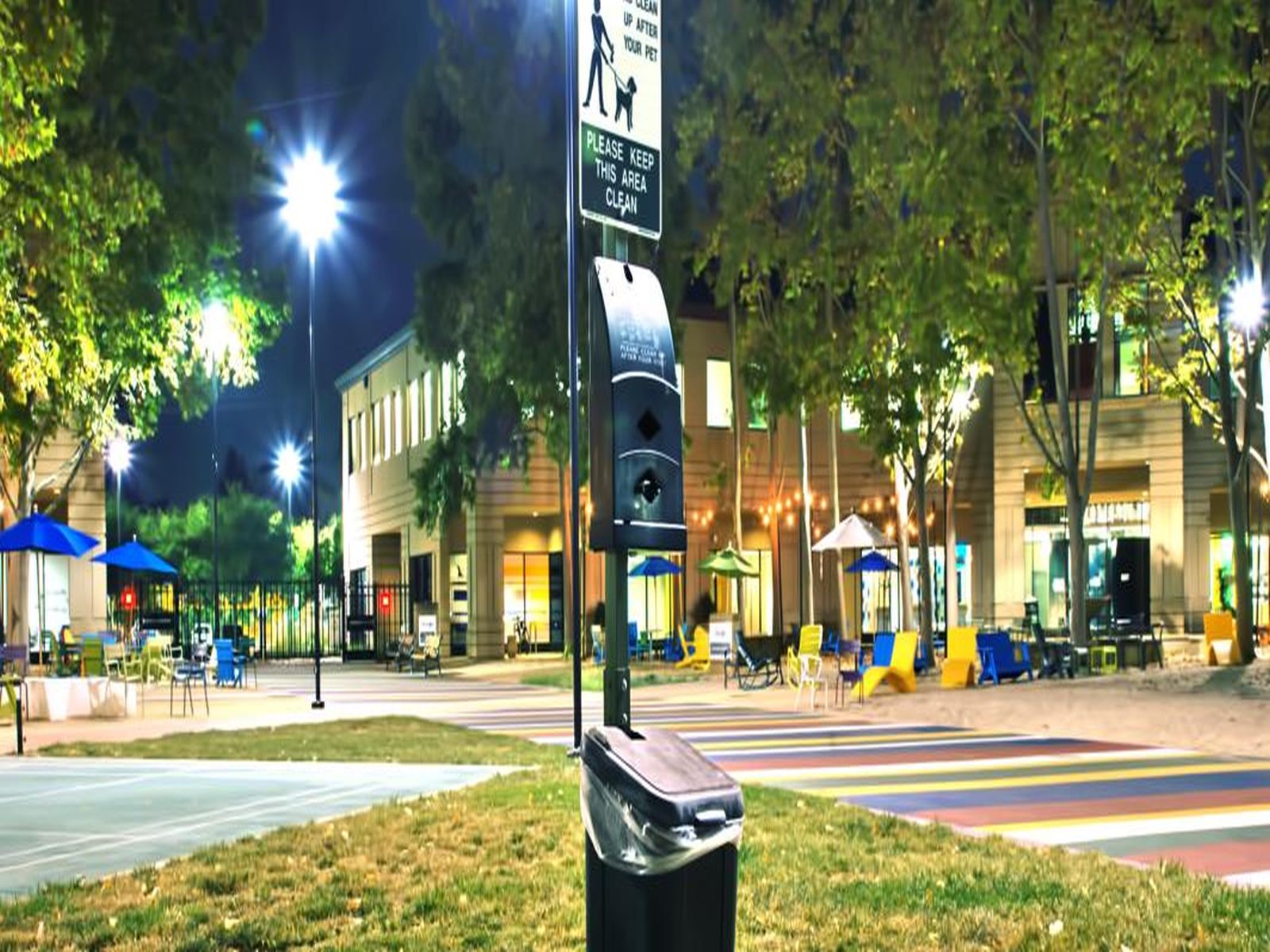}}\hfill
		\subfloat[Our Output (PSNR: \textbf{24.61})]{\includegraphics[width=0.247\textwidth]{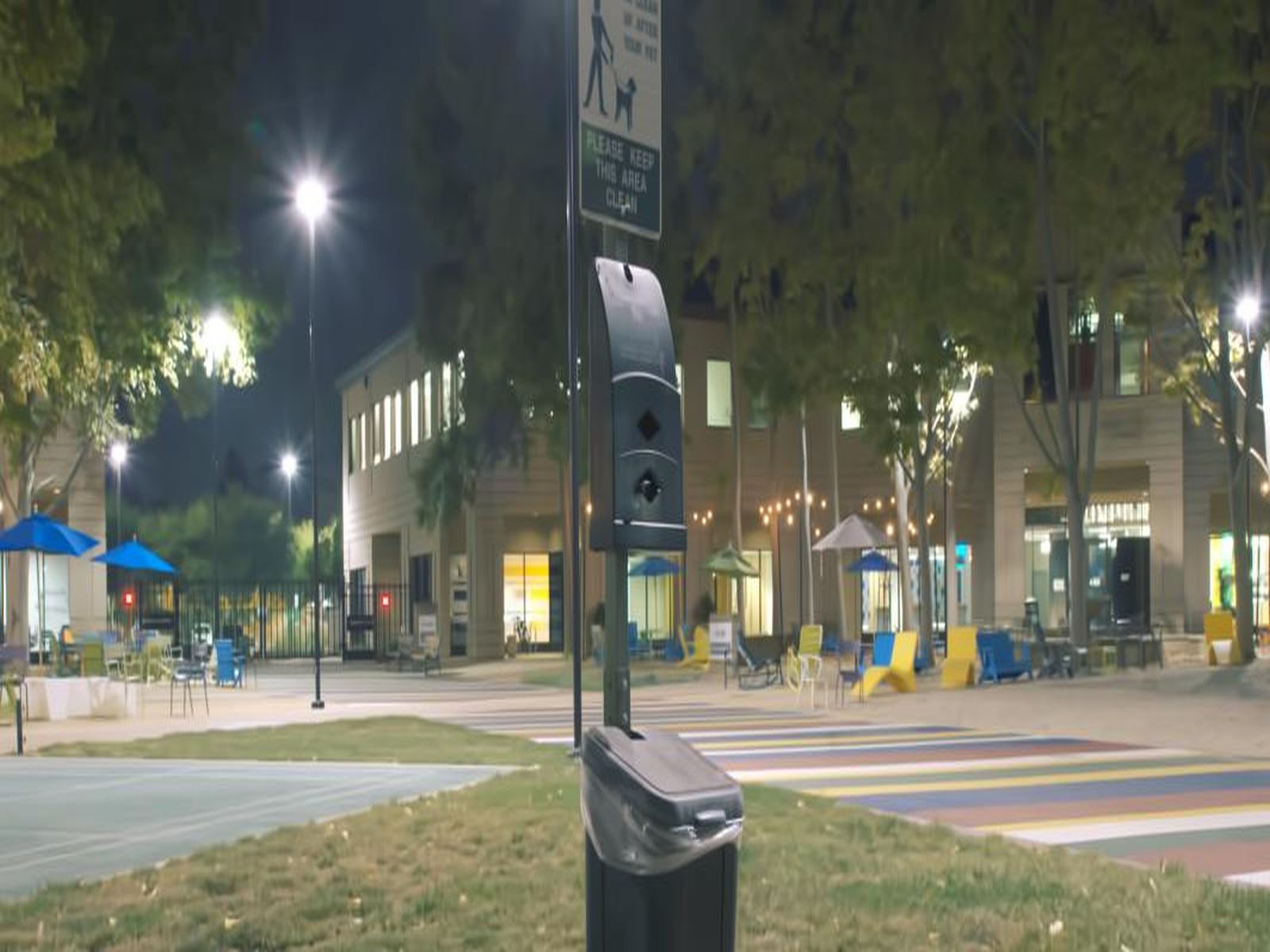}}\hfill
		\subfloat[Ground Truth]{\includegraphics[width=0.247\textwidth]{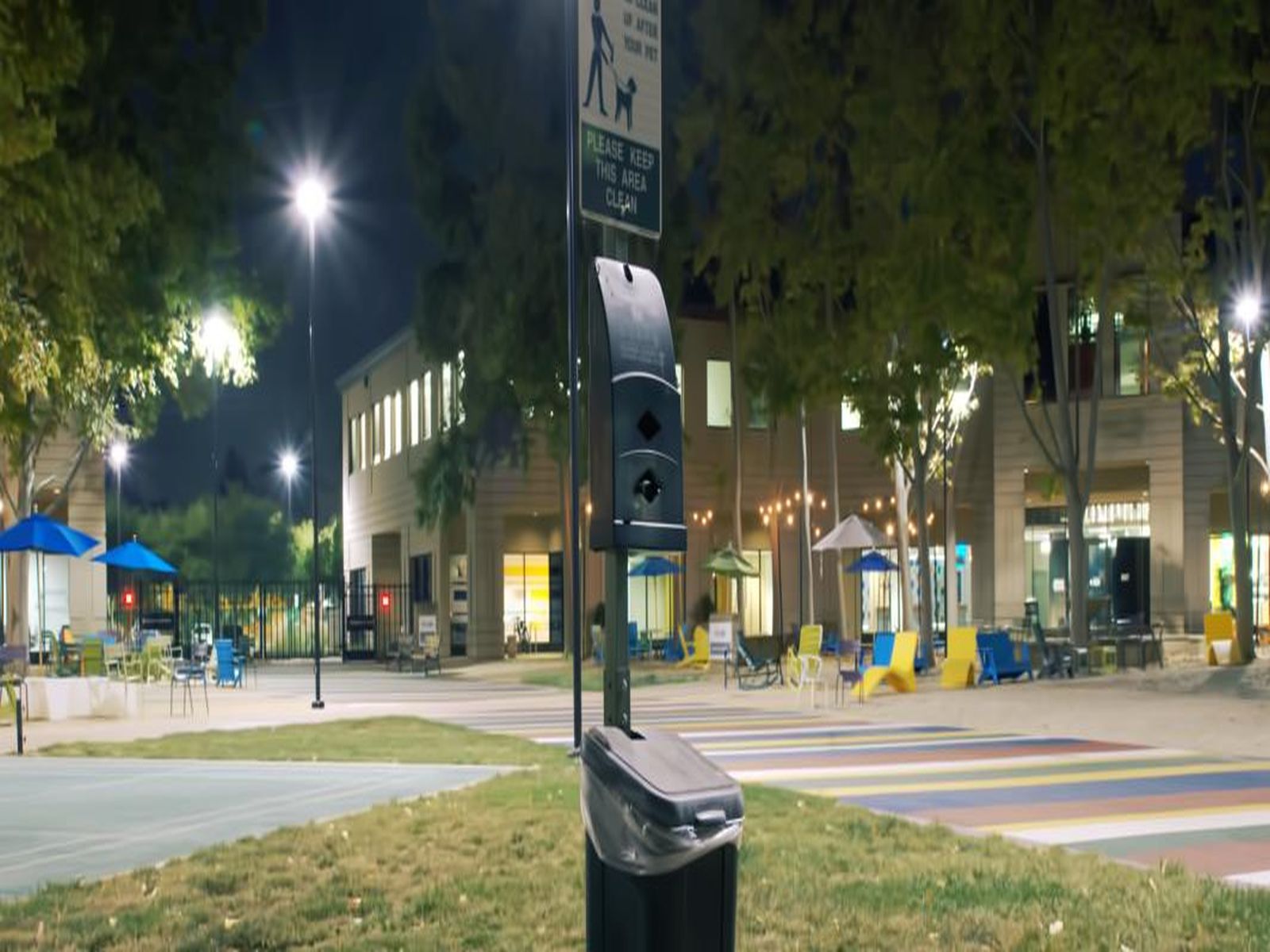}}\hfill
	\end{center}
	\caption{We show an application of our CRF estimation method for nighttime visibility enhancement. As we can observe, our enhanced image is closest to the ground truth image (highest PSNR), showing that our enhancement is more physically correct.}\label{figure_enh_application}
\end{figure*}

\section{Experimental Results}
We first evaluate our non-deep-learning method on general daytime images taken in indoor and outdoor settings. For this, we use the Color-Constancy dataset~\cite{cheng2014illuminant}. We create a test set of 120 images picked randomly from the dataset. The images are taken from three cameras: NikonD40, Canon600D and SonyA57. In the patch selection process, we use the parameters $\{s,\tau_\text{us}, \tau_\text{os}, \tau_\text{un}, \tau_\text{na}\}$ = $\{21, 0.15, 0.9, 0.01,\allowbreak 0.065\}$. The parameter for patch resolution, $s$, is set so that we can have sufficient number of pixels in the pixel distribution to compute the line fitting error. The parameters $\tau_\text{un}$ and $\tau_\text{na}$ are tuned to avoid selecting the uniform patches, and inspecting that the distributions are widely spread in the RGB space. We implement our method in \textsc{Matlab} and use the function \texttt{fmincon} to optimise the coefficients for CRF prediction (Eq.~(\ref{eq_stage_dist_pred})). We set the parameter $\lambda$=0.01 and $\tau_\text{re}$=0.3, that provide the best performance on a separate validation set of 30 images. At stage $t$, we estimate $c=t$ number of coefficients. Wie set $T$=2, as it offers a good tradeoff between our method's accuracy and runtime (as shown in our ablation study). Also, as shown in \cite{ng2007using}, the GGCM model with two coefficients is more accurate than the other models~\cite{grossberg2004modeling,shi2010self,lee2012radiometric} using the same number of coefficients. 

For the baseline methods, we use EdgeCRF~\cite{lin2004radiometric}, CRFNet~\cite{li2017crf} and GICRF~\cite{ng2007using}. Similar to ours, EdgeCRF and GICRF are not learning based methods, while CRFNet is a learning based method. Since the codes of EdgeCRF and CRFNet are not available, these methods are based on our implementation. To evaluate the accuracy of the CRF result from each method, we compute Root Mean Squares Error (RMSE) between the method's CRF result $\hat{g}(\mathrm{x})$ and the ground-truth CRF $g(\mathrm{x})$ by: $\text{RMSE} = \sqrt{\sum_{\mathrm{x}\in \mathrm{x}}(\hat{g}(\mathrm{x})-g(\mathrm{x}))^2}$ (where the ground-truth CRFs are obtained by using images with Macbeth ColorChecker~\cite{lin2004radiometric,lin2005determining,ng2007using,grossberg2004modeling} that are provided by the datasets). For each method, we compute the mean, median, standard deviation, minimum and maximum RMSE values obtained on the entire test set. The quantitative results are shown in Table~\ref{table_daytime_results} and the qualitative CRF results are shown in Fig.~\ref{figure_daynighttime_results}. From the results, we can observe that our CRF results are more accurate and stable than the baseline methods. 

\begin{figure*}[t!]
	\vspace{-0.15in}
	%\captionsetup[subfloat]{labelformat=empty}
	\captionsetup[subfloat]{farskip=1.5pt}
	%\captionsetup[subfloat]{position=top}
	\begin{center}
		\subfloat[Input Image and Detection Results ]{\includegraphics[width=0.48\textwidth]{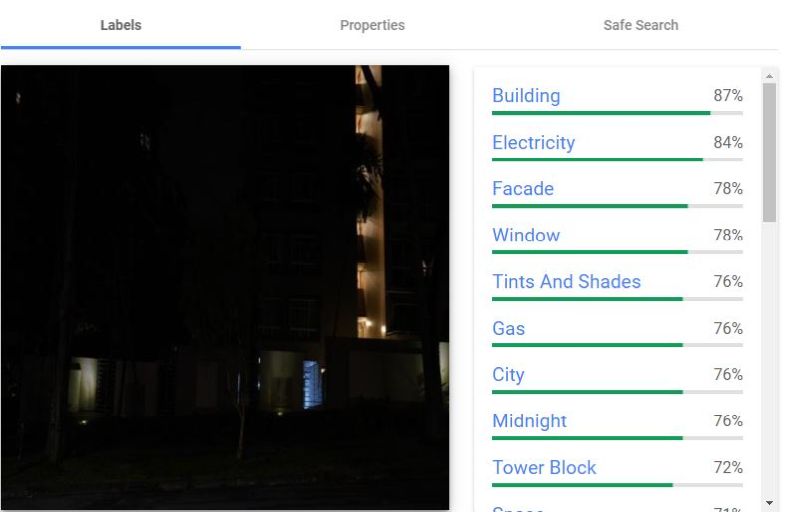}}\hfill
		\subfloat[Our Output and Detection Results ]{\includegraphics[width=0.48\textwidth]{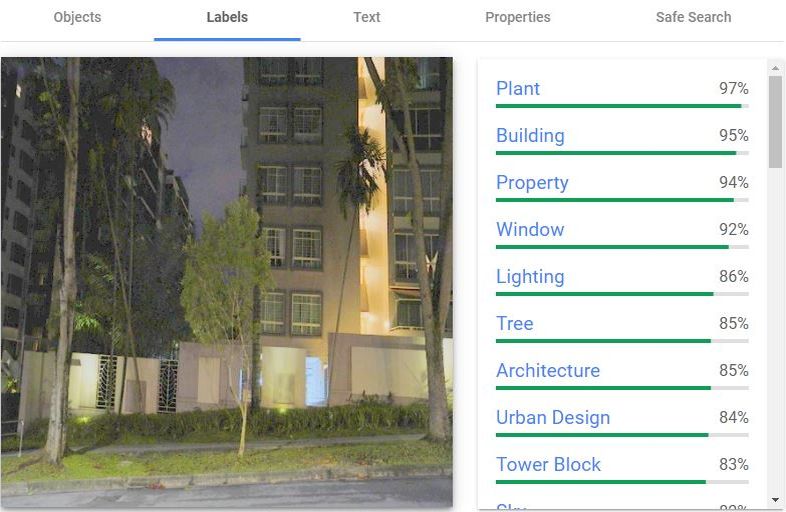}}\hfill
	\end{center}
	\caption{Our visibility enhancement application also helps in generating better results for vision tasks such as object detection in nighttime conditions (e.g. from the results in (b), we can observe more correct predictions for plants, trees, etc., which are completely missed in the results in (a)). Note that, the object detection results are generated using Google Cloud Vision API.}\label{figure_det_application}
\end{figure*}

\begin{figure}[t!]
	\captionsetup[subfloat]{labelformat=empty}
	\captionsetup[subfloat]{farskip=1pt}
	\begin{center}
		\subfloat{\label{figure_ablation1}\includegraphics[width=0.24\textwidth]{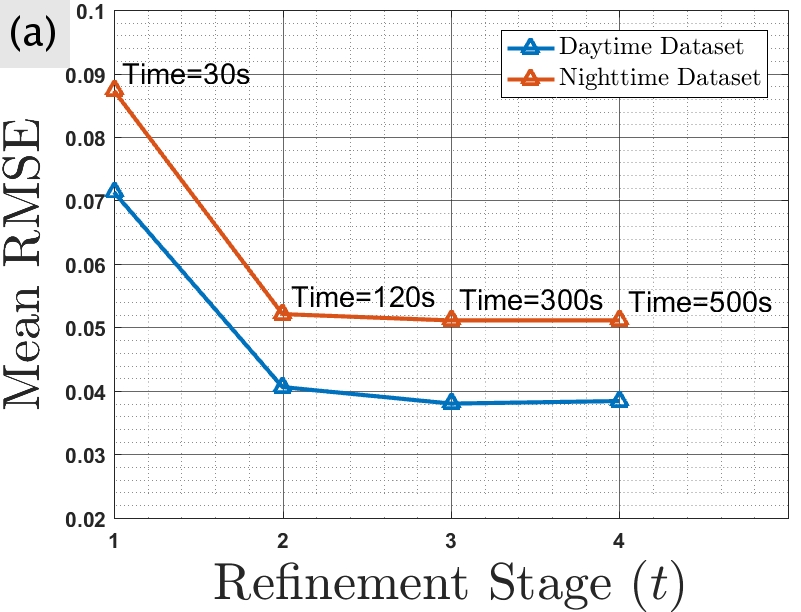}}\hfill
		\subfloat{\label{figure_ablation2}\includegraphics[width=0.24\textwidth]{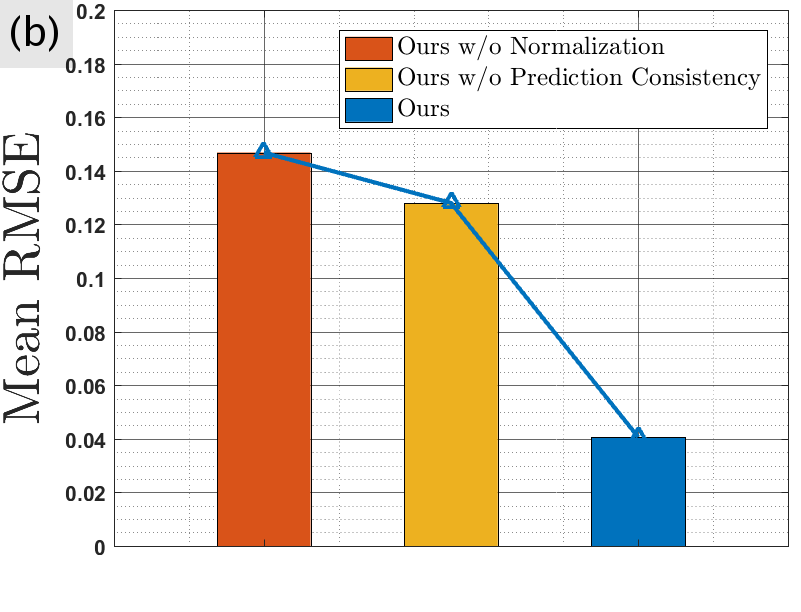}}\hfill\\
	\end{center}
	\caption{(a) Effectiveness of our gradual refinement scheme as the results improve gradually with the refinement stage. (b) Using the normalization operation and the prediction consistency are important for the better performance of our method.}\label{figure_ablation}
\end{figure}

We evaluate our method on nighttime images with varying illumination conditions. For this, we use 20 images taken from our own NikonD80 camera, and 30 images from the SID dataset~\cite{chen2018learning} taken from Sony$\alpha$7s camera. For patch selection, the same parameters used in the daytime experiments are used, except for $\tau_\text{us}$ which is relaxed to 0.02 to allow more patches if the images are low-light. We compare our method with the baseline methods, results corresponding to which are shown in Table~\ref{table_nighttime_results} and Fig.~\ref{figure_daynighttime_results}. We can again observe the better performance of our method compared to the baseline methods. 

The methods of EdgeCRF and GICRF rely on edges and non-locally planar pixels for their respective algorithms, which can be erroneous under general noisy and nighttime conditions. Especially when the images are near low-light, most of the patches are noisy and nearly achromatic, and they cannot be used for CRF estimation (Sec.~\ref{sec_noiseaffect}). In contrast, our method estimates the CRF by taking into account the reliability of the patches in generating the CRF estimate. As observed in Sec.~\ref{sec_noiseaffect}, even if there is a single reliable patch in an image (which can come from the relatively well-illuminated regions in a low-light image), the CRF can be reliably estimated. In addition, contrast to the one-attempt optimisation used by these methods, we use a gradual refinement scheme to gradually improve the CRF results. These factors contribute to our method's better performance for both general daytime and  well-to-dim-lit nighttime images.

The results also show the challenges in using a supervised learning based method such as CRFNet for CRF estimation. CRFNet is trained on the 201 CRFs from the DoRF dataset~\cite{grossberg2004modeling}. If the testing CRFs are the same as the training CRFs, then CRFNet performs better than our method (in terms of mean RMSE, using the same CRF model, CRFNet's score is 0.0201 while ours is 0.0315). However, if the testing CRFs are different (such as the ones used in our experiments), then our method performs better since it is not learning-based.

Our method shows better performance for general real images compared to both learning-based and non-learning-based baseline methods. However, compared to learning-based methods like CRFNet~\cite{li2017crf}, our method has a relatively high computational time ($\sim$120-180 secs on a CPU machine implemented in \textsc{Matlab}). To address this, we propose our deep-learning extension (see Sec.~\ref{sec_dl_extn}), which unlike CRFNet~\cite{li2017crf}, finetunes our CRF estimation network on the test input image in test-time training using unsupervised losses. As shown in Table~\ref{table_nighttime_results2}, it provides better generalisation capability to our method and also improves the CRF estimation time ($\le$4 secs on a GPU machine implemented in \textsc{PyTorch}). Note that, without our proposed test-time training, our CRF estimation performance (both in terms of runtime and CRF estimation accuracy) is similar to that of CRFNet~\cite{li2017crf}.

\begin{table}[!t]
	\centering
	\vspace{0.1in}
	\setlength\belowcaptionskip{-5pt}
	\renewcommand{\arraystretch}{1.2}
	\caption {Comparisons with baselines of our deep learning variant on the daytime test dataset. The numbers represent RMSE. Bold font indicates lowest error} \label{table_nighttime_results2}		
	\begin{tabularx}{\columnwidth}{ c|Y|Y|Y }
		\toprule
		Method & Mean & Min & Max \\
		\midrule
		%\hline
		HDRCNN~\cite{hdrcnn} ($\hat{g}(\mathrm{x}) = \mathrm{x}^{2.2}$) &0.1701 &0.1217 &0.2113\\
		%\hline           
		CRFNet~\cite{li2017crf} &0.1609  &0.0459 &0.2788\\
		%\hline           
		\textbf{Our Method (DL)} &\textbf{0.0902}  &\textbf{0.0384} &\textbf{0.1915}\\
		\bottomrule           
	\end{tabularx}	
\end{table}

\begin{figure*}[t!]
	\captionsetup[subfloat]{labelformat=empty}
	\captionsetup[subfloat]{farskip=1pt}
	\begin{center}
		\subfloat{\includegraphics[width=0.56\textwidth]{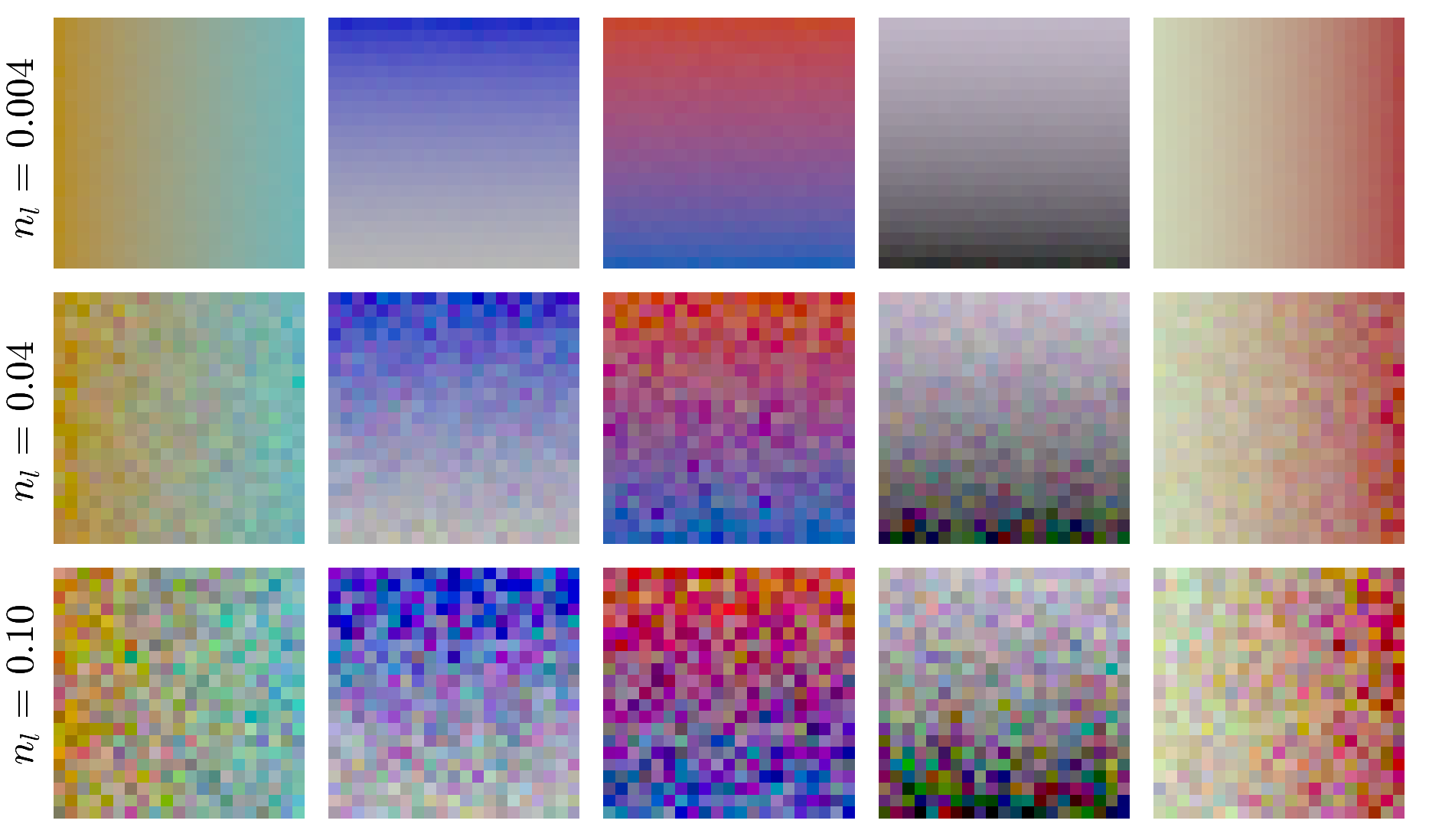}}\hspace{0.1in}
		\subfloat{\includegraphics[width=0.41\textwidth]{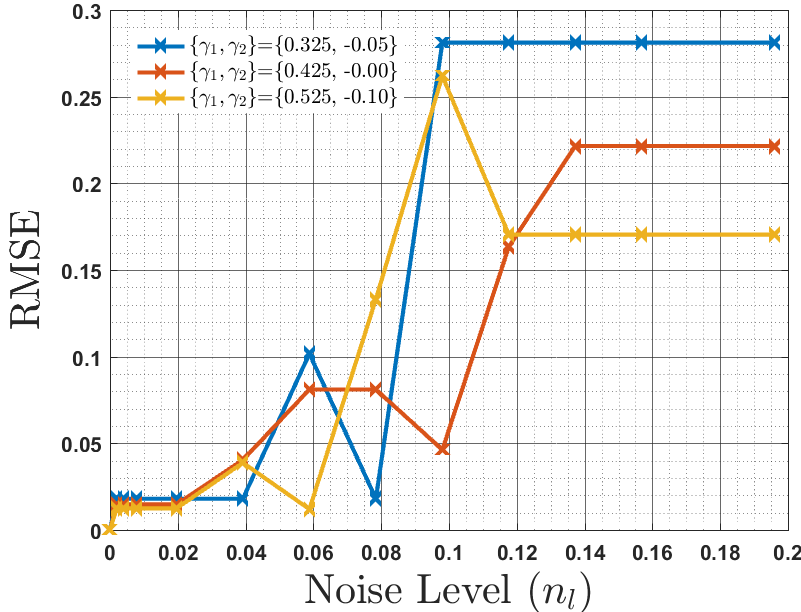}}\hfill\\
		%\vspace{-0.1in}
	\end{center}
	\caption{Performance of our method w.r.t. the noise level $n_l$ (or the standard deviation of noise).  Examples of non-uniform patches generated with varying noise levels and colour mixtures are shown on the left. Our method's performance w.r.t the noise level $n_l$ for three different CRFs is shown on the right.}\label{figure_noise_tolexp1}
	%\vspace{0.1in}
\end{figure*}

\section{Applications}
We show that our CRF estimation method has practical applications, such as visibility enhancement under nighttime conditions. Most existing enhancement methods either assume the CRF to be linear (LIME~\cite{guo2016lime}) or require the RAW image (irradiance image) as the input (SID~\cite{chen2018learning}). The former generates enhancement in a physically incorrect manner, and the latter though being physically correct, is limited in application since RAW images are not available in most practical situations. 
%za
If we can estimate the CRF, we can linearise the intensity image such that it becomes linearly related to the irradiance image. Therefore, we can combine our CRF estimation method with SID to create an enhancement method that accepts an RGB image as input, and enhances the image in a physically correct manner. The results are shown in Fig.~\ref{figure_enh_application}. We can observe that our enhanced image is closest to the ground truth (highest PSNR), thus showing that the enhancement achieved by our method is more physically correct. 

In Fig.~\ref{figure_det_application}, we show another application of our proposed nighttime visibility enhancement. As shown in the results, we can use our method to enhance the visibility of a given input nighttime image and our enhanced output image performs better on vision tasks such as object detection compared to the input nighttime image. To generate the detection results, we use the publicly available Google Cloud Vision API. 

\section{Ablation Study}
\noindent\textbf{\bf{Gradual Refinement}} Fig.~\ref{figure_ablation1} shows that our performance improves gradually with the refinement stage $t$. Since our method's runtime also increases with $t$, we use $T$=2 by default as it offers a good tradeoff between our method's runtime and performance. Also, without gradual refinement (i.e. using one-attempt optimisation), our method's performance drops and the mean RMSE increases to 0.0831 from 0.0406 on the daytime test dataset. 
\smallbreak
\noindent\textbf{Normalization and Prediction Consistency} We compare our method with two variants: (1) ours without using the normalization operation; and (2) ours without using the prediction consistency. The results are shown in Fig.~\ref{figure_ablation2}. We can observe that both the normalization operation and prediction consistency are important factors for the better performance of our method.

\smallbreak
\noindent\textbf{Noise Tolerance} To examine our method's tolerance to noise, we conduct an experiment. The details are as follows. We use 100 synthetic non-uniform patches generated using the GGCM CRF model:  $\mathbf{I} = f(\mathbf{E}) = (\mathbf{E})^{\gamma_1 + ... +\gamma_c\mathbf{E}^{c-1}}$, where the coefficients of the model, $\{\gamma_1, ..., \gamma_c\}$, are set differently for a different trial of the experiment. We add random noise to the patches with varying noise levels. The noise level means the standard deviation of the noise, and is referred by $n_l$. We add random noise independently to the three colour channels of the patches. For each noise level and trial of the experiment, we record the performance of our method in terms of RMSE of the predicted CRF. The results are shown in Fig.~\ref{figure_noise_tolexp1}. 

From Fig.~\ref{figure_noise_tolexp1}, we can observe that our method shows good performance with $\text{RMSE}<0.05$ when noise levels are low to moderate, i.e. when $n_l\le0.04$ (see patches in the middle row of Fig.~\ref{figure_noise_tolexp1}). However, for high noise levels, i.e. $n_l\ge0.10$ (see patches in the bottom row of Fig.~\ref{figure_noise_tolexp1}), our method is unable to estimate the CRF properly. This is because for high noise levels, the linearisation errors of the distributions are highly affected by noise than compared to non-linearity, which makes it difficult to estimate the CRF from the distributions.

\section{Conclusion}
We have presented a method for CRF estimation from a single image using prediction consistency and gradual refinement. Under the presence of noise, we have showed that not every patch is reliable, and it is important to take into account its reliability in estimating the CRF. To handle this problem, we have proposed to use consistency between a patch's CRF predictions as a measure of its reliability. Our method puts more weight on the more reliable patches that provides more accurate results. In addition, we have employed a gradual refinement scheme in the CRF estimation that gradually improves the CRF results. Compared to the existing learning and non-learning-based methods, our method provides more accurate results for both general daytime and nighttime real images. To improve the efficiency and practicality, we further introduced a deep learning extension by proposing a network to estimate the CRF from a single image. Our network also learns from the test input image by performing test-time training on the image using unsupervised losses, which provides our method better generalization performance than the existing learning-based methods.

% or


%\appendices
%% you can choose not to have a title for an appendix
%% if you want by leaving the argument blank
%\section{}

\section*{Acknowledgement}
The authors would like to acknowledge support from MOE2019-T2-1-130. They are thankful to Dr. Lionel Heng, DSO National Laboratories, Singapore, for all the useful discussions and feedback.

\ifCLASSOPTIONcaptionsoff
  \newpage
\fi



%\bibliographystyle{IEEEtran}
\bibliographystyle{IEEEtran}
\bibliography{IEEEabrv,egbib}


\begin{IEEEbiography}[{\includegraphics[width=1in,height=1.25in,clip,keepaspectratio]{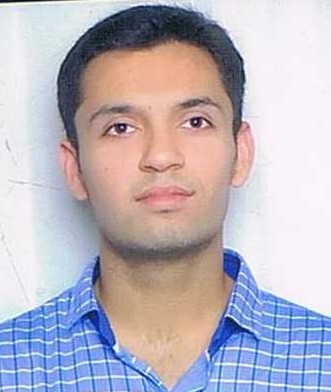}}]{Aashish Sharma}
	Aashish Sharma received his B.Eng. degree in Electronics and Communications Engineering from Delhi College of Engineering, Delhi University in 2012. He is currently pursuing his Ph.D. degree in Computer Vision and Deep Learning at the Electrical and Computer Engineering Department, National University of Singapore, Singapore. His research interests include low-level vision, image enhancement and restoration problems, particularly under bad weather and nighttime conditions.
\end{IEEEbiography}
\vskip -1\baselineskip plus -1fil
\begin{IEEEbiography}[{\includegraphics[width=1in,height=1.25in,clip,keepaspectratio]{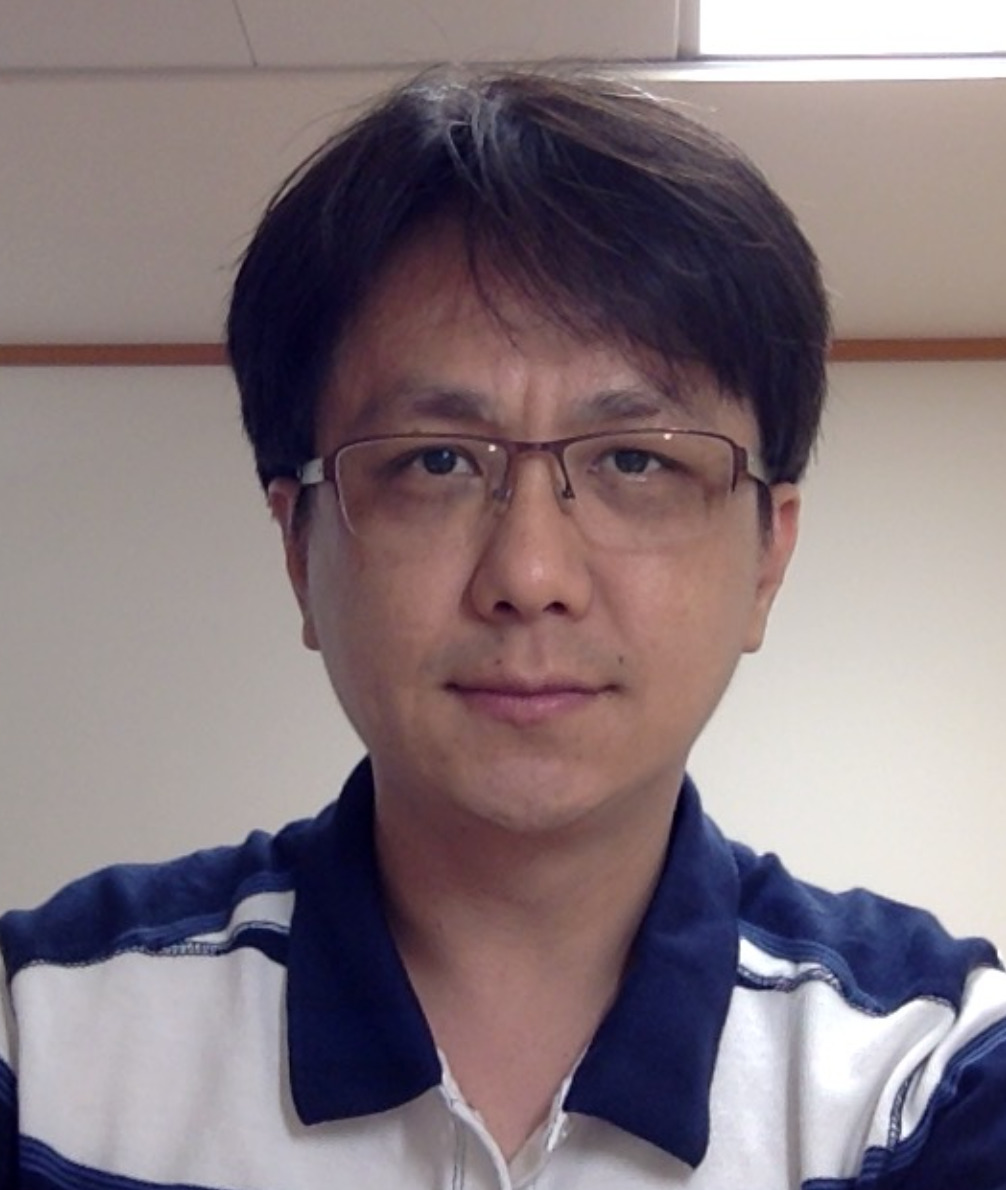}}]{Robby T. Tan}
	Robby T. Tan received the PhD degree in computer science from the University of Tokyo. 
	He is now an associate  professor at both Yale-NUS College and ECE (Electrical and Computing Engineering), National University of Singapore.
	Previously, he was an assistant professor at Utrecht University. 
	His research interests include computer vision and deep learning, particularly in the domains of low level vision (bad weather/nighttime, color analysis, physics-based vision, optical flow, etc.), human pose/motion analysis, and applications of deep learning in healthcare. 
	He is a member of the IEEE.
\end{IEEEbiography}
\vskip -1\baselineskip plus -1fil
\begin{IEEEbiography}[{\includegraphics[width=1in,height=1.25in,clip,keepaspectratio]{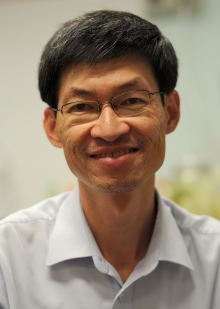}}]{Loong-Fah Cheong}
	Loong-Fah Cheong received the BEng degree from the National University of Singapore, and the PhD degree from the University of Maryland at College Park, Center for Automation Research, in 1990 and 1996, respectively. In
	1996, he joined the Department of Electrical and Computer Engineering, National University of Singapore, where he is an associate professor currently. His research interests include the processes in the perception of three-dimensional motion, shape, and their relationship, as well as	the 3D motion segmentation and the change detection problems.
\end{IEEEbiography}

\end{document}